%% file: harts-arxiv-v2.tex
\documentclass[sigplan,screen,nonacm,balance=false]{harts-arxiv}
\usepackage{asystem-tech-report}
\input{metadata}

\input{preamble}

\title[\PaperTitle]{\PaperTitle}
\input{authors}

\begin{document}
\begin{abstract}
\input{sections/abstract}
\end{abstract}
\maketitle

\input{sections/01_introduction}
\input{sections/02_background_related_work}
\input{sections/03_design}
\input{sections/04_implementation}
\input{sections/05_evaluation}
\input{sections/06_conclusion}

\arxivflushfloats
\appendix
\numberwithin{equation}{section}
\numberwithin{figure}{section}
\numberwithin{table}{section}
\numberwithin{algorithm}{section}
\renewcommand{\theHequation}{appendix.\thesection.\arabic{equation}}
\renewcommand{\theHfigure}{appendix.\thesection.\arabic{figure}}
\renewcommand{\theHtable}{appendix.\thesection.\arabic{table}}
\renewcommand{\theHalgorithm}{appendix.\thesection.\arabic{algorithm}}

\input{appendices/a_scheduling}
\input{appendices/b_minimum_call}
\input{appendices/c_backend_semantics}
\onecolumn
\input{appendices/d_planning_cases}

\twocolumn
\balance
\bibliographystyle{ACM-Reference-Format}
\bibliography{references}
\end{document}

%% file: metadata.tex
\newcommand{\PaperTitle}{HARTS: A Prefix-Sharing Training System for Agentic RL with Hybrid-Attention Models}

%% file: preamble.tex
\usepackage{microtype}
\usepackage{booktabs}
\usepackage{multirow}
\usepackage{amsmath}
\usepackage{algorithm}
\usepackage{algorithmic}
\providecommand{\theHalgorithm}{\arabic{algorithm}}

\newcommand{\harts}{\textsc{Harts}}
\newcommand{\ind}{\mathbb{I}}

\input{layout}

\hypersetup{
  pdftitle={\PaperTitle},
  pdfsubject={HARTS technical report, arXiv v2 working draft},
  pdfkeywords={agentic reinforcement learning, hybrid attention, rollout trees, prefix sharing, distributed training}
}

%% file: layout.tex
\usepackage{balance}

\AtBeginDocument{\raggedbottom}
\makeatletter
\renewcommand{\@mktitle@iii}{%
  \report@originaltitle
  \setbox\mktitle@bx=\vbox{%
    \hsize=\textwidth
    {\raggedright\noindent
      \includegraphics[width=\reportlogowidth]{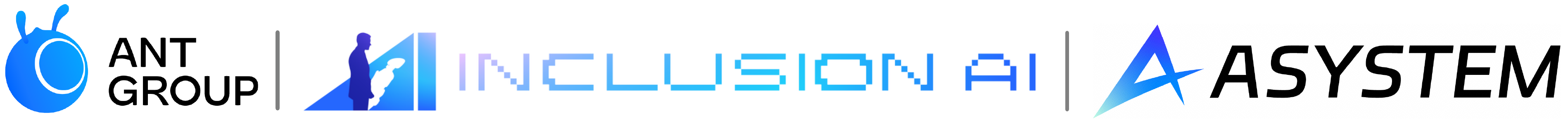}\par}%
    \vskip 12pt
    \reportdividerline
    \vskip 12pt
    \unvbox\mktitle@bx
    \reportdividerline
    \vskip 12pt
  }%
}
\let\arxiv@originallist\@listi
\def\@listi{%
  \arxiv@originallist
  \topsep=5pt\partopsep=0pt\itemsep=5pt\parsep=0pt
}
\let\@listI\@listi
\newcommand{\arxivflushfloats}{%
  \ifx\@deferlist\@empty\else\clearpage\fi
  \ifx\@dbldeferlist\@empty\else\clearpage\fi
}

\patchcmd{\@BAlancecol}
  {\setbox\@leftcolumn=\vbox to \@tempdima}
  {\advance\@tempdima by\dp1\relax
   \setbox\@leftcolumn=\vbox to \@tempdima}
  {}{\PackageError{harts-arxiv-layout}{Cannot adjust reference-column balance}{}}

\let\arxiv@originalauthors\@mkauthors@iii
\let\arxiv@originalabstract\@mkabstract
\renewcommand{\@mkauthors@iii}{%
  \arxiv@originalauthors
  \global\setbox\mktitle@bx=\vbox{%
    \hsize=\textwidth
    \unvbox\mktitle@bx
    \begingroup
      \normalfont\normalsize
      \linewidth=\textwidth
      \parindent=0pt
      \arxiv@originalabstract
    \endgroup
    \vskip 12pt
  }%
}
\let\@mkabstract\@empty
\makeatother

%% file: authors.tex
\reportauthor{Boyuan Meng}{1}
\reportauthor{Peihua Bao}{1}
\reportauthor{Hong Liu}{1}
\reportauthor{Xiaowei Zhu}{1}
\reportauthorbreak
\reportauthor{Chao Wang}{1}
\reportauthor{Gen Li}{1}
\reportauthor{Zhenxuan Pan}{1}
\reportaffiliation{1}{Ant Group, China}
\makeatletter
\g@addto@macro\@authornotes{%
  \begingroup
  \footnotetext{Contact:
    \href{mailto:mengboyuan.mby@antgroup.com}{mengboyuan.mby@antgroup.com}}%
  \endgroup}
\makeatother
\renewcommand{\shortauthors}{Meng et al.}

%% file: sections/abstract.tex
Agentic reinforcement learning (RL) generates irregular rollout trees, but
trajectory-by-trajectory training recomputes their shared interaction histories.
Existing training-time prefix-reuse systems do not jointly address state
recovery, dense execution, and distributed training for hybrid-attention models.
We present \harts{} (Hybrid-Attention RL over Tree Structures), to our knowledge
the first system to demonstrate training acceleration from arbitrary rollout-tree
prefix sharing on real hybrid-attention models.
A two-level planner constructs compression-oriented microbatches, then jointly
refines their membership, cross-microbatch sharing, placement, and execution order.
Within each sharing window, gradients from dependent microbatches accumulate
before one backward pass through the shared prefix computation.
Differentiable boundary-state handoff and bounded replay preserve the
per-trajectory chunkwise numerical path, while a linear-time planner attains
the global minimum number of sequential linear-attention calls under our packed
chunkwise-call model.
The runtime uses a common compact layout across attention and dense or
mixture-of-experts (MoE) layers, with support for distributed parallelism and
activation recomputation.
Recovering log-probabilities for each original trajectory position preserves RL
objectives; for deterministic top-$k$ routing without token dropping, occurrence
counts restore MoE objective weights and expert-load statistics.
On SWE-bench, \harts{} delivers $4.77$--$10.45\times$ the update throughput of
trajectory-expanded training across two sampling schemes on 7.9B- and
124B-parameter hybrid-attention MoE models using 8--64 GPUs.
On the 7.9B model, full-model logit differences are comparable to baseline
self-rerun variation. The 124B model shows stable online learning trends through
step~150 on SWE-bench and $\tau^3$-Bench.

%% file: sections/01_introduction.tex
\section{Introduction}
\label{sec:introduction}

\begin{figure*}[t]
  \centering
  \includegraphics[trim=0 10bp 0 0,clip,width=0.93\textwidth]{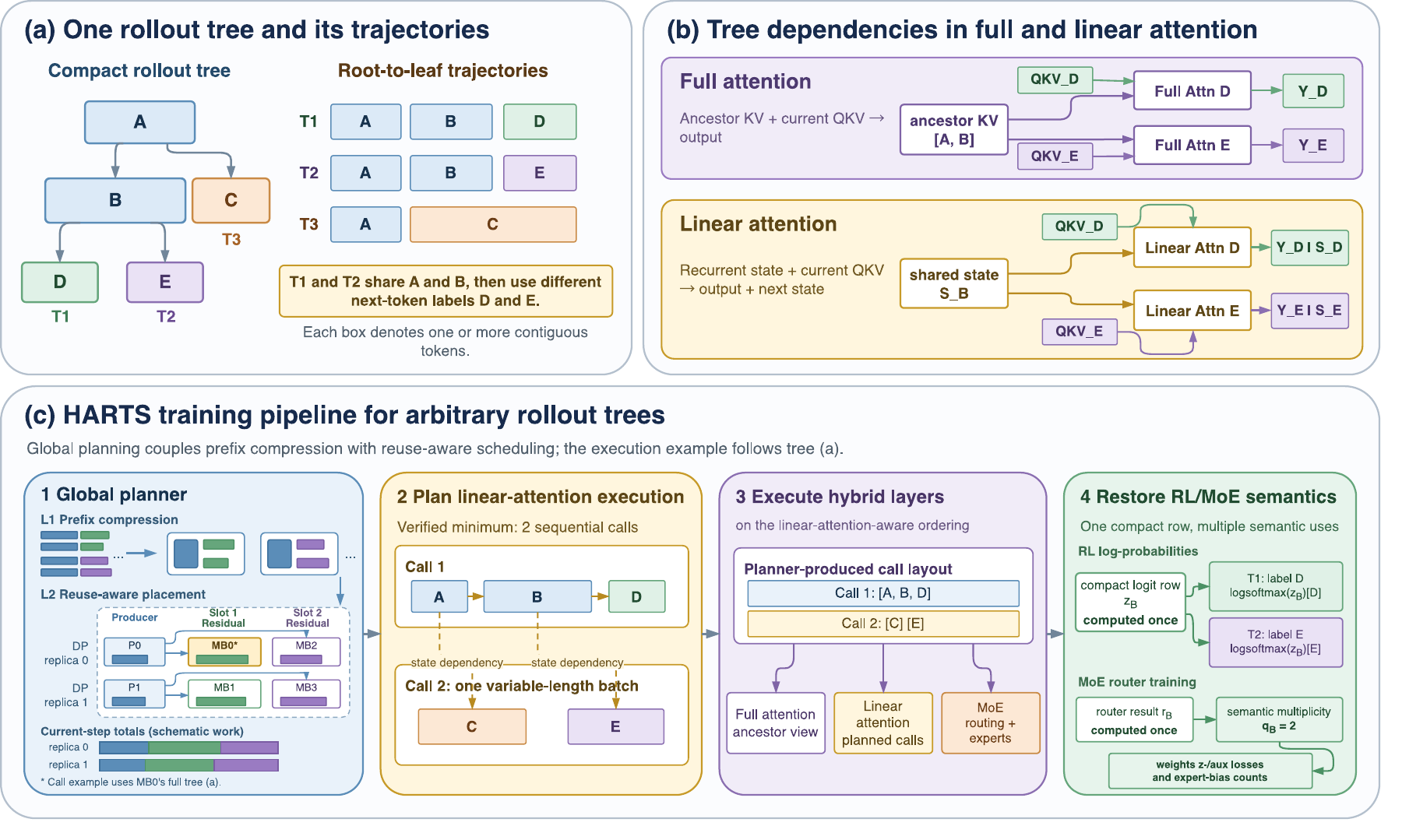}%
  \caption{\textbf{HARTS carries a rollout tree through distributed hybrid-attention training.}
  (a) Three trajectories share token prefixes but may require different next-token
  labels. Boxes represent contiguous token ranges.
  (b) Full attention reads ancestor keys and values; linear attention consumes a recurrent
  state and produces a branch-specific next state.
  (c) L1 constructs compression-oriented microbatches; L2 refines membership,
  sharing, placement, and order. A slot aligns consumers across data-parallel
  replicas: MB0/MB1 form Slot~1 and MB2/MB3 form Slot~2. Producers P0/P1 compute
  shared prefixes for the consumers on their respective replicas. Bars show
  producer and per-slot residual work; stacked bars show current-step totals.
  All bar lengths are schematic. The call and layer examples use MB0's complete
  tree in (a): Call~1 executes $[A,B,D]$, and Call~2 batches independent $[C]$ and
  $[E]$. A common compact ordering supports both attention types and MoE execution.
  Within MB0, one logit row $z_B$ supplies T1/T2's distinct target
  log-probabilities; occurrence count $q_B=2$ restores MoE objective weights and
  expert-load counts without repeating routing or expert computation.}
  \label{fig:harts-overview}
\end{figure*}

Agentic reinforcement learning (RL) trains models through interactions with
environments, tool calls, parallel exploration, and retries. These interactions
can form an irregular \emph{rollout tree}: each root-to-leaf path is a training
trajectory, and trajectories share earlier interaction histories
(Figure~\ref{fig:harts-overview}(a)). For example, \emph{Step-wise RL} uses the
cumulative history through each interaction step as a training sample;
\emph{Tree-Step RL} also branches from shared histories and constructs step-wise
samples along each branch~\citep{yu2026psrl}. Both increase the number of training
samples without proportionally increasing rollout work, but sharing histories
during rollout does not eliminate repeated computation during training.

Trajectory-expanded training repeats forward and backward computation for shared
prefixes within the \mbox{same update}. This cost becomes more prominent as sampling
produces more overlapping histories: across agent workloads, psRL reports that
updates account for $40\%$--$58\%$ of training time with structured sampling,
compared with $19\%$--$32\%$ with trajectory-level
sampling~\citep{yu2026psrl}. Training-time prefix reuse therefore addresses a
growing update-time bottleneck.

Hybrid-attention models make this reuse difficult. A tree mask can express which
ancestor keys and values (KV) a full-attention query may read, but linear attention
also propagates a recurrent state (Figure~\ref{fig:harts-overview}(b)). Efficient
linear-attention kernels advance this state in fixed-length
chunks~\citep{sun2023retnet,yang2024gla,yang2024deltanet,yang2025gateddeltanet,kimiLinear2025}.
An arbitrary tree fork need not coincide with a reusable chunk boundary.
Sharing a prefix thus requires recovering the correct branch state without
changing the numerical path of independent chunkwise training. It also requires
accumulating gradients through shared states and grouping irregular branches
into dense calls, rather than issuing a small kernel for every tree node.

Existing systems study tree-structured training~\citep{wang2026treetraining,zhang2026arealdta}
and cross-microbatch prefix reuse~\citep{li2026scheduleprefixreuse,yu2026psrl},
but do not jointly address these state, numerical, and execution requirements
for arbitrary rollout trees in distributed hybrid-attention training.

We present \harts{} (Hybrid-Attention RL over Tree Structures), a system that
connects prefix compression, reuse-aware scheduling, and dense execution across
hybrid-attention layers (Figure~\ref{fig:harts-overview}(c)). It recovers the
original per-trajectory training interface from compact outputs. To our
knowledge, \harts{} is the first system to demonstrate training acceleration
from arbitrary rollout-tree prefix sharing on real hybrid-attention models.
\harts{} makes three contributions:

\begin{itemize}
\item
We jointly optimize prefix compression and execution balance. A two-level
planner constructs compression-oriented microbatches, then refines
their membership, sharing windows of consecutive microbatches, data-parallel
(DP) replica assignment, and execution order. Placement uses separate costs
for window producers and per-slot \mbox{consumer residuals}.

\item
We combine numerically faithful minimum-call planning with differentiable
execution. Boundary-state handoff and bounded replay preserve the
per-trajectory chunkwise numerical path. A linear-time planner attains the
global minimum number of sequential linear-attention calls under our packed
chunkwise-call model. Shared execution supports distributed parallelism and
activation recomputation, with one producer backward pass per sharing window
after consumer gradients accumulate.

\item
We preserve RL and MoE semantics with compact execution. Compact logits recover
each trajectory position's log-probability, preserving RL
objectives. For deterministic top-$k$ routing without token dropping,
occurrence counts restore mixture-of-experts (MoE) objective weights and
expert-load statistics while routing and expert computation remain compact.
\end{itemize}

We evaluate \harts{} in end-to-end agentic RL runs on
SWE-bench~\citep{jimenez2024swebench}, reporting update throughput. On Tree-Step
RL, it delivers $4.77\times$ baseline throughput with the 7.9B-parameter
Ling-3.0-tiny~\citep{inclusionai2026ling3tiny} on 8 GPUs, and $8.36\times$ and
$8.33\times$ with the 124B-parameter
Ling-3.0-flash~\citep{inclusionai2026ling3flash} on 32 and 64 GPUs, respectively.
On Step-wise RL, Flash achieves
$10.29\times$ and $10.45\times$ on 32 and 64 GPUs. On Tiny, compact execution
provides a further $1.99\times$ throughput improvement over node-wise linear
attention. Separate Tiny experiments show
full-model logit differences comparable to baseline self-rerun variation.
Online Flash training shows stable learning trends through step 150 on SWE-bench and
the banking domain of $\tau^3$-Bench~\citep{shi2026tauknowledge}.

%% file: sections/02_background_related_work.tex
\section{Background and Related Work}
\label{sec:background}

\subsection{Rollout Trees and Compact Representation}

Expanding a rollout tree into independent trajectories gives a set
$\mathcal S$ of \emph{semantic positions}: $s=(x,i)$ denotes token occurrence
$i$ in trajectory $x$, and $S=|\mathcal S|$. Deduplication is defined within each
microbatch. Positions with identical token prefixes share one \emph{compact
token}. A \emph{compact token row} represents that token in a hidden-state or
logit tensor. Let $\mathcal C$ contain the compact tokens across
all microbatches, with $C=|\mathcal C|$. Each training microbatch, called a
\emph{consumer}, retains its own
row identities, even when decoder representations are cached across microbatches.
Sharing model computation does not merge training objectives: positions that
share a row can still have different next-token labels, advantages, masks, or
other RL metadata.

\subsection{Hybrid-Attention Training}

A \emph{hybrid-attention model} interleaves full- and linear-attention layers;
its feed-forward layers may be dense or MoE. Full attention reads ancestor keys
and values through a tree mask or an equivalent view. Linear attention also
propagates a recurrent state, typically using kernels that process fixed-length
execution chunks rather than invoking the recurrence token by
token~\citep{sun2023retnet,yang2024gla,yang2024deltanet,yang2025gateddeltanet,kimiLinear2025}.
Ordinary training interfaces do not expose internal chunk-boundary states for
another branch to consume. For a fork inside a chunk, preserving the independent
trajectory's chunkwise numerical path requires starting from the nearest
preceding boundary and processing the intervening ancestor tokens together with
the branch's new tokens in the same chunkwise call. This \emph{replay} of a fixed
prefix shorter than one chunk \mbox{advances} the state without creating additional
semantic outputs. Reusing states also requires differentiable handoff
between branches.

\subsection{Training-Time Prefix Reuse}

\paragraph{Tree-structured training.}
Tree Training~\citep{wang2026treetraining} and
AReaL-DTA~\citep{zhang2026arealdta} reuse rollout-tree prefixes. Tree Training
evaluates distributed performance, memory, and learning on full-attention
Qwen3-30B/32B~\citep{yang2025qwen3}, not complete hybrid-attention models.
Its linear-attention path executes nodes in small sequential calls and changes
ordinary trajectory training's chunk partitioning. State handoff avoids repeated
ancestor forward passes, but separate backward calls for local leaf losses and
partition cut points still repeatedly traverse shared upstream computation.
Weighting a single unique-token loss cannot recover per-position
log-probabilities for different branch labels. AReaL-DTA targets full-attention
Transformers, limits activation residency through node-wise execution, and
supports data parallelism only. \harts{} instead preserves the per-trajectory
chunkwise numerical path, batches branches into dense hybrid-attention calls,
and executes shared upstream backward once after accumulating gradients.

\paragraph{Cross-microbatch reuse.}
Schedule-Level Shared-Prefix Reuse~\citep{li2026scheduleprefixreuse} and
psRL~\citep{yu2026psrl} cache prefix KV, accumulate consumer gradients, and
execute one producer backward. The former supports multiple parallelism
strategies and multiplicity-weighted deterministic token-local MoE statistics,
but shares only one common prefix per flat group, missing intermediate histories
shared within suffixes. psRL combines group-level placement with token-wise
microbatching, but binds each shared-history group to one worker; local splitting
cannot redistribute its load across workers. Neither KV-centric interface
provides chunkwise boundary-state recovery, numerically aligned replay, and
cross-call planning for arbitrary tree forks. \harts{} instead schedules complete trajectories, not indivisible shared-prefix
groups, and jointly refines microbatch membership, multilevel sharing,
DP-replica assignment, and slot order using separate costs for window producers
and consumer residuals.

\paragraph{Other approaches to prefix redundancy.}
Prefix sharing reduces repeated prompt computation in paired preference
tuning~\citep{wang2024dpoprefixsharing} and in Prefix Grouper's group relative
policy optimization (GRPO) training~\citep{liu2025prefixgrouper}.
DualKV~\citep{gai2026dualkv} combines microbatch repacking with shared-prompt
forward and backward attention kernels.
At the data-processing level, QwenGyre~\citep{wang2026qwengyre} organizes
execution records into prefix-shared trajectory trees. It selects root-to-leaf
training paths and masks previously covered policy targets, counting shared
targets once per execution while retaining their original conditioning contexts.

%% file: sections/03_design.tex
\section{Design: HARTS}
\label{sec:design}

\subsection{Overview}
\label{sec:overview}

\harts{} accepts complete trajectories from arbitrary rollout trees, their
next-token labels, and the original RL metadata. Consumers retain these trajectories
and objectives; a \emph{producer} computes shared prefix decoder representations
for consumers in a sharing window. The system retains a mapping
$m:\mathcal S\rightarrow\mathcal C$ from each semantic position to its
consumer-specific compact token row. Only identical token prefixes share a row;
unrelated trees may share a microbatch but not model computation.

Figure~\ref{fig:harts-overview}(c) connects four stages. The global planner
jointly optimizes prefix compression and reuse-aware load balance. The
minimum-call planner groups independent linear-attention sequences into
execution rounds and fixes their ordering in memory. The runtime executes full
and linear attention and dense/MoE layers over this compact ordering, using
cached producer representations for shared prefixes. Each consumer restores its
complete compact layout for the language-model (LM) head, then recovers current
log-probabilities in the original token order and restores MoE statistics.
Caching does not merge consumers' row identities or training objectives.

\subsection{Joint Scheduling for Prefix Compression and Load Balance}
\label{sec:prefix-microbatch-planning}
\label{sec:placement}

Let $X=\{x_1,\ldots,x_N\}$ be the input trajectories, $D$ the data-parallel
degree, and $T$ the microbatch's compact-token capacity. A trajectory is indivisible: every
$x_i$ belongs to exactly one consumer microbatch. A \emph{DP replica} is a
logical model replica whose computation may span tensor-, sequence-, pipeline-,
and expert-parallel process groups. For $K$ consumer microbatches, each replica receives $K/D$ nonempty
consumers; their corresponding local indices form \emph{microbatch slots}.
Complete consumer layouts and producer forests must satisfy capacity limits.

\noindent\textbf{L1: Prefix compression.}\enspace
For a candidate partition $\mathcal P$, let $w(M)$, its \emph{compact-token work},
count the unique, non-replay compact tokens in microbatch $M$'s complete layout.
Total work sums these counts across microbatches, counting a shared prefix
separately in every microbatch that contains it. For fixed input size $S$,
minimizing total work maximizes within-microbatch prefix compression.
Let $M_{d,j}$ denote replica $d$'s
consumer in slot $j$. L1 compares candidates lexicographically by
\begin{equation}
\begin{aligned}
&J_{\mathrm{init}}(\mathcal P)=\bigl(\sum_{M\in\mathcal P}w(M),\\
&\qquad\sum_j\max_d w(M_{d,j}),\;\max_d\sum_j w(M_{d,j})\bigr).
\end{aligned}
\label{eq:prefix-plan-objective}
\end{equation}
The latter two terms break compression ties using the sum of per-slot maximum
loads and the maximum total load on any replica. Exact trie and longest-common-prefix (LCP) costs guide
capacity checks and cuts. For each specified candidate count, small inputs use
an exact minimum-work search over contiguous intervals in token-lexicographic
order; large inputs use
sampled-count merge/split search. An additional trie-aware greedy construction
explores noncontiguous membership at either scale. Initial assignment groups
similarly sized microbatches into slots and assigns heavier work to less-loaded
replicas. Appendix~\ref*{supp:scheduling-l1} gives the searches and their scope.

\begin{figure*}[t]
  \centering
  \includegraphics[trim=0 46bp 0 0,clip,width=\textwidth]{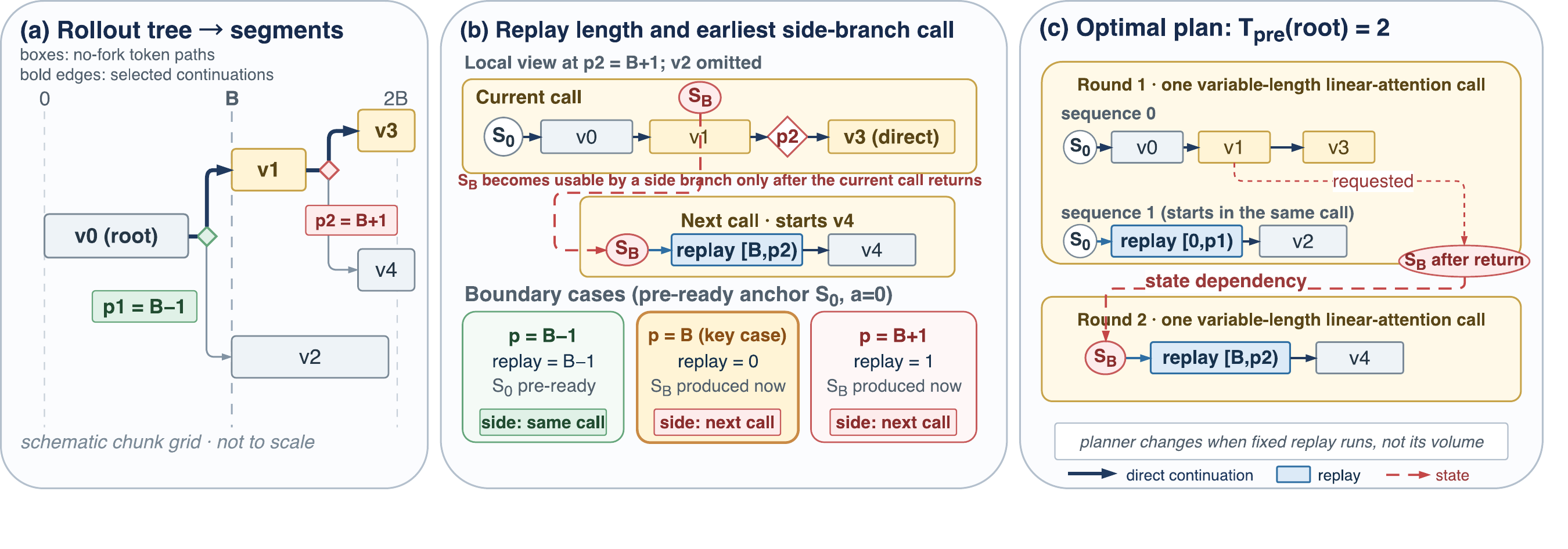}%
  \caption{\textbf{From tree forks to minimum-call linear-attention execution.}
  (a) Boxes are fork-free segments; thick edges select direct continuations
  $v_1$ and $v_3$. (b) At $p_2$, the state $\mathbf S_B$ produced by the current
  call can initialize the independent side branch only after that call returns.
  The three boundary cases distinguish replay length from earliest legal start;
  this local view omits $v_2$. (c) Round~1 packs $v_0\!\rightarrow v_1\!\rightarrow
  v_3$ with replay-started $v_2$; Round~2 executes replay-started $v_4$.
  All sequences in each round form one packed call.}
  \label{fig:minimum-call}
\end{figure*}

\noindent\textbf{L2: Reuse-aware placement.}\enspace
L2 assigns L1's microbatches to DP replicas and slots, balancing work at
corresponding slots while placing prefix-affine microbatches in consecutive
sharing windows. These decisions are coupled: placement determines sharing
opportunities, and sharing changes each consumer's execution work.

All replicas use the same slot indices and window boundaries. Within each
replica, a \emph{producer forest} merges and deduplicates selected trajectory
prefixes at multiple branching levels. Different branches can serve different
subsets of consumers, rather than only computing a prefix common to every
microbatch in the window. Consumers reuse their matching branches and run
the decoder only on the remaining tokens. In Figure~\ref{fig:harts-overview}(c), P0/P1 each
serve one replica's consumers across Slots~1--2. Producers add no consumer slots;
consumers retain complete compact layouts for the LM head and objectives.

L2 evaluates work after sharing. For plan $\Pi$, let $\mathcal J$ index consumer
slots and $\mathcal W(\Pi)$ its sharing windows. Estimated forward/backward
costs for replica $d$ are $\widehat c^{\mathrm{F,C}}_{d,j}$,
$\widehat c^{\mathrm{B,C}}_{d,j}$ for the consumer in slot $j$, and
$\widehat c^{\mathrm{F,P}}_{d,w}$, $\widehat c^{\mathrm{B,P}}_{d,w}$ for the
producer of window $w$. These estimates combine token counts including
hardware-alignment padding with full-attention query/key sequence lengths,
including ancestor keys. L2 targets the bottleneck cost
\begin{equation}
\begin{aligned}
\Phi(\Pi)={}&\sum_{j\in\mathcal J}
\left(\max_d\widehat c^{\mathrm{F,C}}_{d,j}
+\max_d\widehat c^{\mathrm{B,C}}_{d,j}\right)\\
&+\sum_{w\in\mathcal W(\Pi)}
\left(\max_d\widehat c^{\mathrm{F,P}}_{d,w}
+\max_d\widehat c^{\mathrm{B,P}}_{d,w}\right).
\end{aligned}
\label{eq:phase-placement-objective}
\end{equation}
Shared-prefix decoder work is charged to the producer, leaving only uncached
work in each consumer's cost. Equation~\eqref{eq:phase-placement-objective}
takes separate maxima across DP replicas for the forward and backward costs
of each producer and consumer slot. This captures their individual bottlenecks
even when replicas have equal total work.

A bounded heuristic explores placements, windows, and sharing depths, with
local whole-trajectory repacking and exchanges that preserve L1's consumer
count. Feasible plans are compared lexicographically by
$\Phi$, then the total producer/residual token count including alignment padding,
normalized DP load imbalance, and maximum total work on any replica. Appendix~\ref*{supp:scheduling-l2}
details the search and cost calculation.

\subsection{Minimum-Call Planning for Linear-Attention Execution}
\label{sec:call-planning}

\paragraph{Execution model.}
Contract each maximal fork-free token path into a \emph{segment}, giving a rooted
tree $G=(V,E)$ with segment $v=[s_v,e_v)$. Segments follow tree forks, not the
backend's chunk grid. A packed call processes independent input sequences,
each potentially spanning multiple chunks. A child may directly continue its
parent's sequence, but a state produced inside the call can initialize another
independent sequence only after the call returns. At most one child per fork
can be a direct continuation. One packed call is an \emph{execution round};
the planner minimizes sequential call depth for each linear-attention layer
of a microbatch.

\paragraph{Fixed replay and state readiness.}
Let fork position $p$ be the shared-prefix length, an exclusive token boundary.
For chunk size $B$, the nearest legal resume anchor is
\begin{equation}
a(p)=B\left\lfloor p/B\right\rfloor.
\label{eq:resume-anchor}
\end{equation}
Write $a_v=a(s_v)$ and $b_v=a(e_v)$. A non-direct child of $v$ starts from
$\mathbf S_{b_v}$ and replays $[b_v,e_v)$ in the linear-attention core. Replay
is fixed, shorter than $B$, and produces no semantic outputs.
If $\mathbf S_{a_v}$ exists before the current round and
\begin{equation}
e_v<a_v+B,
\label{eq:early-start}
\end{equation}
then $b_v=a_v$, so side children can start in the same round from that ready
state. Otherwise their anchor is produced inside the current call, requiring
another round. The inequality is strict: at $e_v=a_v+B$, replay is zero but the
new state is still unavailable to an independent sequence until the next call.
Figure~\ref{fig:minimum-call}(b) contrasts forks at $B-1$, $B$, and $B+1$.

\paragraph{Minimum-call recurrence.}
Let $T_{\mathrm{pre}}(v)$ be the minimum rounds to finish $v$'s subtree when
$\mathbf S_{a_v}$ is ready before the current call. Let $T_{\mathrm{in}}(v)$
be the corresponding minimum when $v$ directly continues its parent and its
entry anchor is produced within the current call. Both include the current round.
A leaf needs one call even if it
spans multiple chunks: $T_{\mathrm{pre}}(v)=T_{\mathrm{in}}(v)=1$.
For a nonleaf, write $\operatorname{ch}(v)$ for its children and define the
maximum over an empty set as zero. Selecting direct child $o$ gives
\begin{equation}
\begin{aligned}
T_{\mathrm{in}}(v)&=\min_{o\in\operatorname{ch}(v)}
\max\Bigl(T_{\mathrm{in}}(o),\\
&\qquad 1+\max_{u\in\operatorname{ch}(v)\setminus\{o\}}
T_{\mathrm{pre}}(u)\Bigr).
\end{aligned}
\label{eq:tin}
\end{equation}
The direct subtree continues without waiting; all side subtrees wait one round
and then run in parallel. With an already-ready entry anchor,
\begin{equation}
T_{\mathrm{pre}}(v)=
\begin{cases}
\displaystyle\max_{u\in\operatorname{ch}(v)}T_{\mathrm{pre}}(u),
&e_v<a_v+B,\\[3pt]
T_{\mathrm{in}}(v),&e_v\geq a_v+B.
\end{cases}
\label{eq:tpre}
\end{equation}
The first case starts all children from the same ready anchor in the current
round. The root's zero state is ready before Round~1, so the result is
$T_{\mathrm{pre}}(\mathrm{root})$. Independent trees in the same microbatch share round-wise packed
calls; their combined minimum is the maximum of their results, not the sum.

In Figure~\ref{fig:minimum-call}, the subtree rooted at $v_1$ needs two rounds:
one leaf continues
directly, while the other waits for $\mathbf S_B$. The ready zero anchor lets
$v_2$ start in Round~1, so the whole tree also needs two rounds.

A postorder pass computes these values; budget-aware recovery chooses direct
continuations and assigns side \mbox{sequences} to rounds. Using the largest and
second-largest child values avoids repeated sibling scans. Planning and
recovery take $O(|V|+|E|)$ time and $O(|V|)$ space. They attain the
\emph{global minimum number of sequential linear-attention calls} under this packed-call
model: replay starts at the nearest legal anchor, each non-replay output is
produced once, and newly produced states initialize independent sequences only
after return. The model does not reduce calls by earlier-anchor multichunk
replay or duplicated segment computation. Mandatory replay is unchanged by
continuation choice. Appendix~\ref*{supp:minimum-call} gives recovery and
replay accounting; Appendix~\ref*{supp:planning-cases} gives full cases.

\paragraph{Optimality.}
Induct on subtree height. For a fixed direct child $o$, the two terms in
Equation~\eqref{eq:tin} are necessary lower bounds, attained by continuing $o$
now and launching all side subtrees together next round. When
Equation~\eqref{eq:early-start} holds with a ready entry anchor, all child
anchors are ready, so their maximum is likewise necessary and attainable.
These cases establish both recurrences from the leaf base case and preserve
$T_{\mathrm{pre}}(v)\leq T_{\mathrm{in}}(v)\leq T_{\mathrm{pre}}(v)+1$.
In particular, let $M=\max_u T_{\mathrm{pre}}(u)$. Choosing a child attaining
$M$ as the direct continuation costs at most $M+1$ rounds; any different
choice leaves that child on a side branch and requires at least $M+1$.
Thus this greedy choice is safe, and the root value is globally minimal under
the stated packed-call model.

\subsection{Differentiable Shared Execution}
\label{sec:hybrid-execution}

The linear-attention call plan (Section~\ref{sec:call-planning}) defines the
compact-token ordering in memory for each decoder computation. Full-attention and
dense/MoE layers use the same layout. Attention may assemble temporary
ancestor-KV, replay, or local-context inputs, but returns compact-token outputs
to their corresponding rows. This avoids reconstructing trajectory-expanded
tensors between layers.

\paragraph{Differentiable state handoff.}
For input sequence $i$ spanning $n_i$ chunks, let $\mathbf Q_i$, $\mathbf K_i$,
and $\mathbf V_i$ denote queries, keys, and values, and $\mathbf O_i$ the outputs.
Let $\mathcal I_i$ index the boundary states requested by the plan. The training operator exposes
\begin{equation}
\begin{aligned}
&\operatorname{ChunkLA}(\mathbf Q_i,\mathbf K_i,\mathbf V_i,
\mathbf S^{(i)}_{[0]};\mathcal I_i)\\
&\quad\rightarrow
\left(\mathbf O_i,\mathbf S^{(i)}_{[n_i]},
\{\mathbf S^{(i)}_{[c]}\}_{c\in\mathcal I_i}\right).
\end{aligned}
\label{eq:chunkla-interface}
\end{equation}
Requested states are exported during the existing chunkwise recurrence, without
a second state scan. Their consumers' initial-state gradients accumulate into
the computations that produced them. Sequences without explicit sharing remain
state- and gradient-isolated. This contract requires a chunkwise operator with
recoverable recurrent states and initial-state support, not a particular update
formula.

Each nonempty round is one packed call. Replay reuses projected inputs and any
causal-convolution results; it repeats only core state updates, not projection,
convolution, dense/MoE computation, or semantic outputs. Bounded local operators
use true ancestor context, not adjacent packed sequences. Full attention gathers
projected ancestor KV into temporary branch views; gather backward sums their
gradients into the shared compact activations.

\paragraph{Window-scoped reuse and backward.}
\label{sec:cross-mb-runtime}
A producer forward caches shared decoder hidden states and per-layer
interfaces: ancestor KV, legal boundary states, replay inputs, router logits,
and required local context. Consumers run the decoder on uncached tokens, then restore
their own complete layouts for the LM head and objectives. Producers execute
no LM head or independent policy loss. Each consumer restores router logits and
uses occurrence counts to compute router losses and expert-load counts;
the producer adds no separate consumer statistics.

Consumer backward passes accumulate interface gradients into the producer graph.
After all dependent contributions arrive, one producer backward pass traverses
the shared upstream computation, then the window's graph, caches, and gradient
buffers are released. Full-layer activation recomputation rebuilds the same
packed inputs, rounds, replay, boundary states, and graph connections from the
fixed plan, without replanning. It preserves window-scoped gradient accumulation
and the single producer backward rule.

\begin{figure}[!tbp]
  \centering
  \includegraphics[trim=0 8bp 0 0,clip,width=\columnwidth]{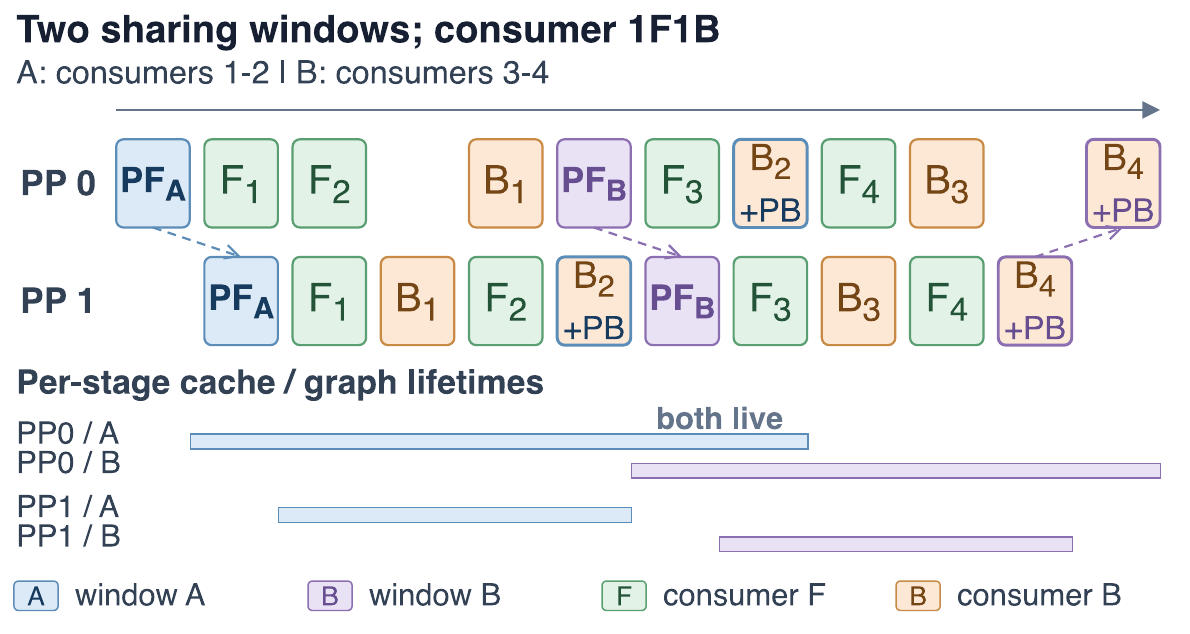}%
  \caption{\textbf{Overlapping sharing windows under consumer 1F1B.}
  A contains consumers 1--2; B contains 3--4. PF/PB denote producer
  forward/backward; $F_i/B_i$ denote consumer forward/backward. PB runs inside
  the window's last consumer backward. Thin bars show stage-local retention
  from PF completion to release after PB; only stage~0 retains both windows
  concurrently. Dashed arrows show selected pipeline dependencies, not global
  barriers; event columns are not measured durations.}
  \label{fig:cross-mb-pipeline}
\end{figure}

\paragraph{Pipeline execution.}
Figure~\ref{fig:cross-mb-pipeline} preserves consumer one-forward-one-backward
(1F1B) order under pipeline parallelism (PP). Producer forward (PF) precedes
the window's first consumer but advances no consumer-forward count and occupies
no consumer slot. Producer backward (PB) runs inside its last consumer backward,
without a separate scheduling slot. The next window's PF may precede the
current window's PB. The planner limits how many producers retain activations
for backward at once; each pipeline stage releases its resources after its
own PB. Appendix~\ref*{supp:backend-semantics}
details parallel execution and backend interfaces.

\subsection{Preserving RL and MoE Semantics without Expansion}
\label{sec:semantic-restoration}

Sharing computation must preserve trajectory-specific targets and weights
without expanding vocabulary-sized \mbox{tensors} or expert execution.

\paragraph{Compact log-probability recovery.}
For vocabulary $\mathcal V=\{1,\ldots,V\}$, let $z_c\in\mathbb R^V$ be compact
row $c$'s logits. Let $\mathcal P\subseteq\mathcal S$ contain nonfinal semantic
positions, $y_s$ be position $s$'s next-token target, and
$\mathcal A\subseteq\mathcal P$ be positions selected by the original action mask.
For every $s\in\mathcal P$,
\begin{equation}
\begin{aligned}
\lambda_s&=\log\operatorname{softmax}(z_{m(s)})_{y_s}\\
&=z_{m(s),y_s}-\log\sum_{v\in\mathcal V}\exp z_{m(s),v}.
\end{aligned}
\label{eq:mapped-logp}
\end{equation}
The implementation computes maxima and sums of exponentials once per compact row, then
looks up each position's own target in original batch order. Different branch
targets can therefore use the same logit row. The original action mask selects
$\boldsymbol\lambda=(\lambda_s)_{s\in\mathcal A}$ only after this restoration.

If the original policy objective accesses model outputs only through these
current next-token log-probabilities and its original metadata
$\boldsymbol\xi$, it remains
$\mathcal L_{\mathrm{policy}}=F(\boldsymbol\lambda,\boldsymbol\xi)$.
Advantages, old/reference log-probabilities, masks, loss weights, trajectory
boundaries, and reduction denominators retain their original organization;
cross-token and cross-trajectory coupling in $F$ is unchanged. Backward sums
each position's own gradient contribution into its compact row, rather than
multiplying one gradient by a reuse count. Forward and backward both take
$O(CV+S)$ work without $S\times V$ logits or gradients.

\paragraph{MoE weights and load statistics.}
Define semantic multiplicity
$q_c=|\{s\in\mathcal S:m(s)=c\}|$, so $\sum_{c\in\mathcal C}q_c=S$.
Under deterministic top-$k$ routing without token dropping, a compact hidden
state undergoes routing, dispatch, and expert computation once. Multiplicity
weights its router z-loss~\citep{zoph2022stmoe}. Weighted routing probability mass and top-$k$
assignment counts form the load-balancing auxiliary loss. Expert-bias updates~\citep{wang2024lossfree} use the same
weighted assignment counts. These weights restore training semantics without
changing routing decisions or repeating expert work.
Appendix~\ref*{supp:backend-semantics} gives the
compact-gradient and vocabulary-parallel details.

%% file: sections/04_implementation.tex
\section{Implementation}
\label{sec:implementation}

\harts{} integrates AReaL~\citep{fu2025areal} with the Megatron training
stack~\citep{shoeybi2019megatron,narayanan2021megatron,korthikanti2023recomputation}. AReaL builds rollout trees and retains token
labels and RL metadata. It runs the two-level global planner and the
minimum-call planner, then materializes producer/consumer views, cache mappings,
rounds, replay intervals, and state sources. Megatron consumes these fixed plans
without traversing the trees or rescheduling during execution.

The concrete backends are Multi-head Latent Attention (MLA)~\citep{deepseek2024v2}, Kimi Delta
Attention (KDA), and dense/MoE feed-forward networks (FFNs) in Ling-3.0-tiny and
Ling-3.0-flash. MLA uses Transformer Engine's asymmetric packed Q/KV views
with original position IDs and bottom-right causal
alignment~\citep{nvidia2026transformerengine}. KDA exports requested boundary
states during its existing recurrence and accepts initial-state gradients in
backward. Replay gathers existing projected/convolved activations and repeats
only the state-update core. Megatron's MoE path keeps dispatch, communication,
and expert computation compact; its other extensions implement mapped
log-probabilities, window-scoped gradient accumulation, and activation recomputation.

Our RL runs use group relative policy optimization
(GRPO)~\citep{shao2024deepseekmath}. Megatron returns current log-probabilities
in original token order; AReaL retains the original group-relative advantages,
old/reference log-probabilities, masks, and reductions. The same interface
supports proximal policy optimization (PPO)~\citep{schulman2017ppo} and other objectives satisfying
Section~\ref{sec:semantic-restoration}'s input contract, without changing their
loss definitions. Appendix~\ref*{supp:backend-semantics} details the
backend and parallel-execution interfaces.

%% file: sections/05_evaluation.tex
\section{Evaluation}
\label{sec:evaluation}

We evaluate update throughput, two-level scheduling, execution ablations,
numerical fidelity, and online RL learning behavior.

\subsection{Experimental Setup}
\label{sec:setup}

\paragraph{Models and workloads.}
We use Ling-3.0-tiny~\citep{inclusionai2026ling3tiny} and
Ling-3.0-flash~\citep{inclusionai2026ling3flash}, two hybrid-attention MoE
models (Table~\ref{tab:model}). Tiny has 7.9B total and 1.3B active parameters
per token; Flash has 124B total and 5.1B active. Both interleave KDA and MLA
within one forward pass and use MoE in most layers. Update-throughput
measurements come from end-to-end agentic RL using a Claude Code scaffold on
SWE-bench tasks~\citep{jimenez2024swebench}.

\begin{table}[t]
\caption{Evaluated hybrid-attention MoE models. Each repeated attention group
contains the listed KDA:MLA ratio.}
\label{tab:model}
\centering
\small
\begin{tabular}{@{}lcc@{}}
\toprule
Property & Tiny & Flash \\
\midrule
Total / active parameters (B) & 7.9 / 1.3 & 124 / 5.1 \\
Transformer layers & 24 & 42 \\
KDA:MLA per group & 3:1 & 5:1 \\
Dense FFN layers & 1 & 1--2 \\
MoE layers & 2--24 & 3--42 \\
Routed experts / top-$k$ & 128 / 8 & 512 / 8 \\
\bottomrule
\end{tabular}
\end{table}

\paragraph{Configurations and baseline.}
Table~\ref{tab:topology} lists \mbox{data-,} \mbox{tensor-,} \mbox{pipeline-,} and
expert-parallel degrees (DP/\allowbreak TP/\allowbreak PP/\allowbreak EP) and microbatch capacities. Performance
runs use H20-3e GPUs, bfloat16 (BF16), and full-layer activation recomputation.
RL uses partial rollout with staleness set to~1. Dropout is disabled in both
systems for all training experiments. Baseline expands trajectories
independently; \harts{} uses global planning, compact execution, and sharing
windows. Comparisons match models, inputs, and parallel configurations without
changing training semantics.

\paragraph{Throughput measurement.}
Each step's update throughput is original trajectory tokens across all DP
replicas divided by full-update wall-clock time, including planning and excluding
rollout. Shared token occurrences are counted separately in each original
trajectory. We report arithmetic means over all post-warmup training steps and
speedups relative to Baseline's mean.

\begin{table}[t]
\caption{SWE-bench update-throughput configurations.}
\label{tab:topology}
\centering
\small
\setlength{\tabcolsep}{3pt}
\begin{tabular}{@{}lrrrrr@{}}
\toprule
Model / GPUs & DP & TP & PP & EP & Tokens/MB cap \\
\midrule
Tiny / 8 & 2 & 2 & 2 & 4 & 65,536 \\
Flash / 32 & 2 & 8 & 2 & 16 & 110,592 \\
Flash / 64 & 4 & 8 & 2 & 16 & 110,592 \\
\bottomrule
\end{tabular}
\end{table}

\subsection{Update Throughput across Models and Parallel Configurations}
\label{sec:throughput-evaluation}

\begin{figure}[t]
  \centering
  \includegraphics[trim=0 8bp 0 0,clip,width=\columnwidth]{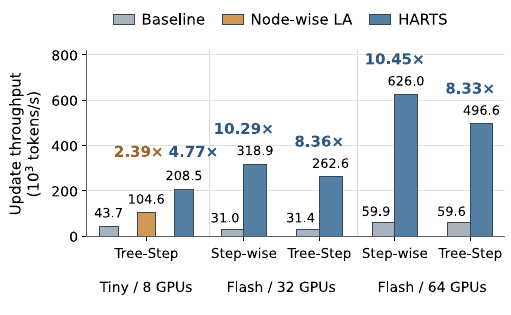}%
  \caption{\textbf{SWE-bench update throughput.} Comparisons use matched inputs;
  speedups are relative to Baseline. Tiny additionally includes Node-wise LA.}
  \label{fig:update-throughput}
\end{figure}

Figure~\ref{fig:update-throughput} shows Tree-Step RL throughput increasing
from 43,745 to 208,492 tokens/s on Tiny/8 ($4.77\times$), from 31,414 to
262,560 on Flash/32 ($8.36\times$), and from 59,612 to 496,616 on Flash/64
($8.33\times$). Step-wise RL increases from 30,993 to 318,893 tokens/s on
Flash/32 ($10.29\times$) and from 59,922 to 625,990 on Flash/64
($10.45\times$). Thus prefix sharing translates into update-throughput gains
across both sampling schemes, models, and device scales.

\begin{table}[t]
\caption{Within-microbatch prefix compression on SWE-bench.}
\label{tab:performance}
\centering
\small
\begin{tabular}{@{}llc@{}}
\toprule
Model / GPUs & Sampling & Compression \\
\midrule
Tiny / 8 & Tree-Step & $3.91\times$ \\
Flash / 32 & Step-wise & $7.58\times$ \\
Flash / 32 & Tree-Step & $6.65\times$ \\
Flash / 64 & Step-wise & $7.57\times$ \\
Flash / 64 & Tree-Step & $6.56\times$ \\
\bottomrule
\end{tabular}
\end{table}

Table~\ref{tab:performance} reports within-microbatch prefix compression:
original trajectory tokens divided by compact tokens summed over complete
microbatch layouts, excluding replay. This is not an upper bound on update
speedup: cross-microbatch reuse further reduces decoder work, and scheduling
shortens synchronization-critical work. Neither benefit is captured by
complete-layout token counts alone.

\subsection{Two-Level Scheduling: Quality and Overhead}
\label{sec:planner-evaluation}

\paragraph{Workload and comparison.}
We replay all ten historical steps (0--9), comprising 16,170 trajectory records
and 435,034,818 original tokens, with $D=8$ and capacity $T=65,536$,
separately from the GPU-throughput runs.
OR-Tools+FFD~\citep{perron2025ortools} first assigns shards by original-token
first-fit decreasing (FFD), partitions prefixes within each DP replica, then aligns
microbatch counts. OR-Tools quality results use a 300-second-per-rank search
budget; planning-time measurements stop search at the first feasible solution.

\paragraph{Metrics.}
Compression is the ratio of original trajectory tokens to non-replay tokens
processed by the decoder after prefix sharing, aggregated over the ten steps.
L1 uses the prefix-compression plan from Section~\ref{sec:placement}; L1+L2
adds reuse-aware placement, jointly optimizing cross-microbatch prefix sharing
and load balance. For DP replica $d$, let $r_{d,j}$ count non-replay tokens processed
by the decoder in microbatch slot $j$, excluding cached shared prefixes, and let
$p_{d,h}$ count shared-prefix tokens processed once for window $h$.
Table~\ref{tab:planner-quality} reports aggregate DP load balance:
\[
\eta_{\mathrm{sched}}=
\frac{\sum_j\sum_d r_{d,j}+\sum_h\sum_d p_{d,h}}
{D\left(\sum_j\max_d r_{d,j}+\sum_h\max_d p_{d,h}\right)}.
\]
Here $d$ ranges over the $D$ replicas, and $j$ and $h$ cover all slots and
sharing windows in the ten steps. Without cross-microbatch sharing,
$r_{d,j}$ counts the complete microbatch and the $p$ terms are absent.
Higher is better; $100\%$ means equal replica loads for every computation.

\begin{table}[t]
\caption{\textbf{Two-level scheduling.}}
\label{tab:planner-quality}
\label{tab:planner-stages}
\label{tab:planner-balance}
\centering
\small
\setlength{\tabcolsep}{3pt}
\begin{tabular}{@{}lccc@{}}
\toprule
Planner & Compression & Balance & \shortstack{Time\\(s/step)} \\
\midrule
OR-Tools+FFD & $5.535\times$ & $94.11\%$ & 3.1705 \\
\harts{} L1 & $5.588\times$ & $98.62\%$ & 0.5097 \\
\harts{} L1+L2 & $7.543\times$ & $97.59\%$ & 0.8891 \\
\bottomrule
\end{tabular}
\par\smallskip
\parbox{\columnwidth}{\raggedright
OR-Tools: compression/balance use a 300\,s/rank search budget;
time measures complete planning with first-feasible stopping.}
\end{table}

L2 further reduces decoder work by $25.91\%$ relative to L1
(Table~\ref{tab:planner-quality}). Balance changes from $98.62\%$
to $97.59\%$ as cross-microbatch sharing redistributes execution work. Against OR-Tools
with the 300-second-per-rank budget,
the final plan improves compression by $36.26\%$ and balance by 3.48 percentage
points: greater reuse and balanced execution are achieved together.

\paragraph{Load balance across DP scales.}
Table~\ref{tab:planner-balance-distribution} complements the overall balance
in Table~\ref{tab:planner-quality} at DP degrees 2, 4, and 8, using the same input
and capacity. For each slot or shared-prefix computation, balance is mean
replica load divided by maximum load. P5 is the fifth percentile of these values,
highlighting poorly balanced computations rather than a single extreme case.
For step-total balance, we sum each replica's non-replay token work within a
training step, then divide the mean of these totals by their maximum.
We report the minimum over ten steps to capture the least-balanced step.
By jointly planning sharing, placement, and order,
\harts{} improves both metrics with fewer microbatch slots. At DP8, P5 and
worst step-total balance improve by 20.69 and 4.51 percentage points,
respectively, while slots decrease from 166 to 152.

\begin{table}[t]
\caption{\textbf{DP load-balance distributions.}}
\label{tab:planner-balance-distribution}
\centering
\small
\setlength{\tabcolsep}{3pt}
\begin{tabular}{@{}cccc@{}}
\toprule
\multirow{2}{*}{DP} & \multicolumn{3}{c}{\harts{} L1+L2 / OR-Tools} \\
\cmidrule(l){2-4}
& \shortstack{Microbatch\\slots} & \shortstack{Balance\\P5 (\%)} &
\shortstack{Step-total\\minimum (\%)} \\
\midrule
2 & $599/614$ & $96.65/87.91$ & $98.40/96.35$ \\
4 & $301/319$ & $94.40/79.62$ & $98.20/95.08$ \\
8 & $152/166$ & $93.61/72.91$ & $95.96/91.45$ \\
\bottomrule
\end{tabular}
\par\smallskip
\parbox{\columnwidth}{\raggedright
OR-Tools: all columns use a 300\,s/rank search budget.
Slots are per-replica totals over ten steps.}
\end{table}

\subsection{Execution Ablations}
\label{sec:execution-ablations}

\paragraph{Compact execution.}
On Tiny/8, \emph{Node-wise LA} executes linear attention per rollout-tree
node's token segment, passing recurrent states along parent--child
dependencies~\citep{wang2026treetraining}. All other components, including
planning and cross-microbatch reuse, remain unchanged. It reaches 104,642
tokens/s ($2.39\times$ Baseline). Batching these calls into dense compact
execution yields another $1.99\times$, reaching 208,492 tokens/s
(Figure~\ref{fig:update-throughput}).

\begin{figure}[t]
  \centering
  \includegraphics[trim=0 3bp 0 0,clip,width=\columnwidth]{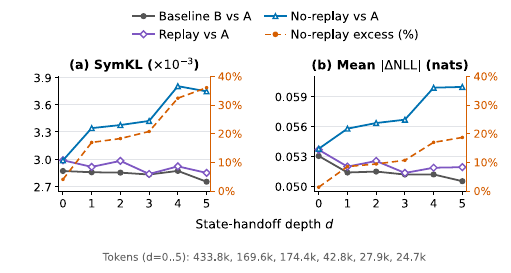}%
  \caption{\textbf{Replay preserves chunkwise numerical fidelity.}
  Solid lines: (a) symmetric KL and (b) mean absolute target-NLL difference,
  each relative to Baseline A. Orange dashed lines: No-replay's excess over
  Baseline B's error, $(E_N/E_B-1)\times100\%$. Token counts appear below
  handoff depths 0--5; lower is better.}
  \label{fig:replay-fidelity}
\end{figure}

\begin{table}[ht]
\caption{PP scheduling ablation on Tiny/8 Tree-Step.}
\label{tab:pp-scheduling}
\centering
\small
\begin{tabular}{@{}lr@{}}
\toprule
PP schedule & Throughput (k tokens/s) \\
\midrule
Non-overlapping windows & 167.7 \\
Overlap + backward-first & 190.4 \\
\harts{} (consumer 1F1B) & 208.5 \\
\bottomrule
\end{tabular}
\end{table}

\begin{table*}[!t]
\caption{\textbf{Full-model fidelity on Tiny.} Cosine is the equally weighted
mean of token-wise full-vocabulary logit cosines. Self-rerun compares baseline
repeats; both \harts{} paths use the matched baseline.
Absolute loss difference uses \harts{} without recomputation.}
\label{tab:cosine}
\centering
\small
\begin{tabular}{@{}lcccc@{}}
\toprule
Parallel configuration & \shortstack{Baseline\\self-rerun} &
\shortstack{\harts{} w/o\\recomputation} &
\shortstack{\harts{} w/\\recomputation} &
\shortstack{Absolute\\loss difference} \\
\midrule
DP8/TP1/PP1/EP8 & 0.99978662 & 0.99977463 & 0.99977816 & $9.66\times10^{-5}$ \\
DP4/TP2/SP/PP1/EP8 & 0.99975827 & 0.99974703 & 0.99974416 & $6.02\times10^{-6}$ \\
DP2/TP2/SP/PP2/EP4 & 0.99972408 & 0.99971610 & 0.99971732 & $1.33\times10^{-4}$ \\
\bottomrule
\end{tabular}
\end{table*}

\begin{figure*}[!t]
  \centering
  \begin{minipage}[t]{0.49\textwidth}
    \centering
    \includegraphics[width=\linewidth]{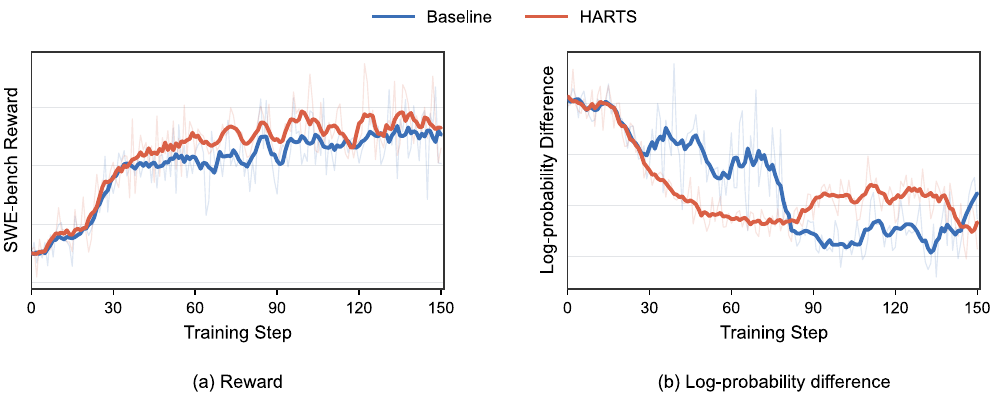}%
  \end{minipage}\hfill
  \begin{minipage}[t]{0.49\textwidth}
    \centering
    \includegraphics[width=\linewidth]{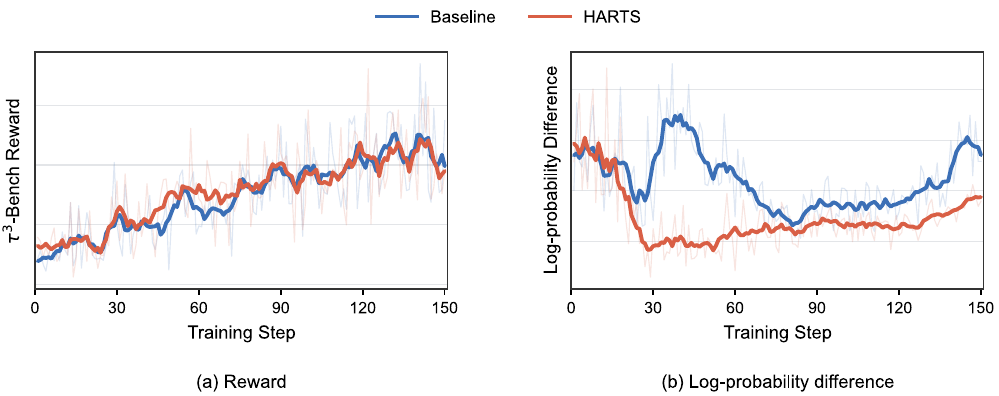}%
  \end{minipage}
  \caption{\textbf{Online Flash training through step 150.}
  Left pair: SWE-bench; right pair: $\tau^3$-Bench.
  Each pair shows (a) reward and (b) the absolute difference between rollout/old and
  training-recomputed log-probabilities within each run.}
  \label{fig:online-rl}
  \label{fig:swe-online}
  \label{fig:reward}
\end{figure*}

\paragraph{Pipeline scheduling.}
On Tiny/8 Tree-Step, only the PP schedule varies
(Table~\ref{tab:pp-scheduling}). The non-overlapping variant uses the PP
schedule of \citet{li2026scheduleprefixreuse} within each window, running one
window at a time. The other two variants overlap
windows: one gives backward work priority among dependency-ready events;
\harts{} preserves consumer 1F1B order across windows
(Figure~\ref{fig:cross-mb-pipeline}).
\harts{} improves throughput by $24.3\%$ over non-overlapping windows and
$9.5\%$ over backward-first scheduling, confirming consumer 1F1B's benefit.

\paragraph{Replay ablation.}
Using Tiny with a fixed checkpoint, rollout tree, and microbatch boundaries,
we compare Baseline A, its repeat Baseline B, \harts{} Replay, and No-replay.
No-replay passes the parent final state without preserving trajectory chunk
boundaries. Relative to A, we report mean token-wise symmetric
Kullback--Leibler (KL) divergence over full-vocabulary distributions and mean
absolute target-token negative-log-likelihood (NLL) difference. The latter
isolates current log-probabilities from advantages, masks, and loss reduction.
Handoff depth $d$ counts logical parent--child segment boundaries, not
materialized state tensors; root tokens have $d=0$.

Replay closely tracks baseline-repeat variation across $d=0,\ldots,5$
(Figure~\ref{fig:replay-fidelity}), whereas No-replay diverges after the first
handoff, with deviations generally increasing with depth. At $d=4$--$5$, its
symmetric KL exceeds baseline-repeat error by $32.4\%$--$36.1\%$, and mean
$|\Delta\mathrm{NLL}|$ by $17.0\%$--$18.7\%$. Nearest-boundary recovery and
less-than-one-chunk replay are therefore necessary to preserve ordinary
chunkwise training's numerical path.

\raggedbottom
\subsection{Training Fidelity and RL Effectiveness}
\label{sec:fidelity-evaluation}

\paragraph{Full-model logits and loss.}
Table~\ref{tab:cosine} covers three parallel configurations, including sequence
parallelism (SP). Both \harts{} paths exceed 0.9997 mean token-wise cosine,
closely matching baseline self-rerun variation. Recomputation under the fixed
plan adds no notable deviation. Absolute scalar-loss differences
($6.02\times10^{-6}$--$1.33\times10^{-4}$) corroborate training fidelity.

\paragraph{Online learning.}
Figure~\ref{fig:online-rl} compares Baseline and \harts{} on Flash through
step 150 on SWE-bench and the banking domain of
$\tau^3$-Bench~\citep{shi2026tauknowledge}.
On $\tau^3$-Bench, both start from the same checkpoint and training
configuration. Faint curves are raw per-step values; thick curves use
seven-step centered moving averages.

On SWE-bench, rewards rise rapidly then level off; \harts{}'s smoothed reward
is slightly higher in most stages. Log-probability differences decrease overall
but change relative order. On $\tau^3$-Bench, rewards rise similarly, with lower
log-probability differences for \harts{} in most stages. Together with numerical
fidelity, these results show stable learning trends in the complete agentic RL
pipeline.

%% file: sections/06_conclusion.tex
\section{Conclusion}
\label{sec:conclusion}

\harts{} turns shared histories in arbitrary rollout trees into efficient
hybrid-attention RL training. Two-level scheduling combines prefix compression
with reuse-aware load balancing. Differentiable compact execution preserves
per-trajectory training semantics across distributed parallelism and
recomputation. A linear-time planner attains the global minimum number of
sequential linear-attention calls under the stated packed chunkwise-call model;
each sharing window executes one producer backward after accumulating consumer
gradients.

On SWE-bench Tree-Step RL, \harts{} delivers $4.77\times$, $8.36\times$, and
$8.33\times$ baseline update throughput on Tiny/8, Flash/32, and Flash/64.
On Tiny, compact execution provides a further $1.99\times$ improvement over
node-wise linear attention.
Tiny's logit differences are comparable to baseline self-rerun variation, with
mean token-wise full-vocabulary cosine above 0.9997. Online Flash training on
SWE-bench and $\tau^3$-Bench shows stable learning trends through step 150.

%% file: appendices/a_scheduling.tex
\section{Scheduling Search Details}
\label{supp:scheduling}

This section expands the two-level planner in
Section~\ref{sec:placement}. Complete trajectories remain
indivisible, and $w(M)=C(M)$ counts non-replay compact tokens in a microbatch's
complete layout. Cross-microbatch caching does not change this count.

\subsection{L1: Prefix Compression}
\label{supp:scheduling-l1}

\paragraph{Exact prefix costs.}

Let $\operatorname{LCP}(x,y)$ be the length of the longest common token prefix.
For a microbatch with trajectories in stable lexicographic order,
\begin{equation}
C(M)=|x_1|+\sum_{i=2}^{k}
\bigl(|x_i|-\operatorname{LCP}(x_{i-1},x_i)\bigr).
\label{eq:compact-trie-cost}
\end{equation}
The immediately preceding trajectory supplies the longest prefix shared with
the trajectories already inserted. Let
$L_i=\operatorname{LCP}(x_{i-1},x_i)$ and let $W(a,b)$ be the compact work of
$x_a,\ldots,x_b$. Cutting before $x_t$ duplicates exactly the boundary LCP:
\begin{equation}
W(a,t-1)+W(t,b)-W(a,b)=L_t.
\label{eq:cut-prefix-loss}
\end{equation}
For any pair of trajectories, their LCP is the minimum adjacent LCP between
them in lexicographic order. Prefix-cost preprocessing and a range-minimum-query
index therefore provide constant-time interval-work and pair-LCP queries.
Capacity and candidate comparisons use the exact compact-token count,
not raw trajectory lengths.

\paragraph{Natural partition and candidate counts.}

The planner recursively splits only intervals whose compact work exceeds $T$.
It cuts at the smallest adjacent LCP, preferring the smaller maximum child work
on ties, then the smaller difference between child works. The first \mbox{feasible}
partition is the \emph{natural partition} $\mathcal P_0$, with $K_0$ microbatches.
For $G=C(X)$, the necessary count bound is
\begin{equation}
K_{\min}=D\max\left(1,\left\lceil\frac{G}{T\cdot D}\right\rceil\right).
\label{eq:k-lower-bound}
\end{equation}
A count is DP-compatible when divisible by $D$. The base search range is
\begin{equation}
\begin{aligned}
K&\in\{K_{\min},K_{\min}+D,\ldots,K_{\max}\},\\
K_{\max}&=\max\left(K_{\min},\min\left(
D\left\lceil\frac{K_0}{D}\right\rceil,
D\left\lfloor\frac{N}{D}\right\rfloor\right)\right).
\end{aligned}
\label{eq:k-search-range}
\end{equation}
$K_{\min}$ does not guarantee feasible groups. Compression is not monotone in
$K$: forcing fewer groups can split long-prefix trajectory clusters. The
planner therefore compares multiple counts rather than choosing the smallest
feasible count.

\paragraph{Contiguous exact search and scalable search.}

For every candidate $K$, exact mode finds the minimum-work nonempty,
capacity-feasible lexicographically contiguous partition. With $F[k,e]$ the
minimum work for partitioning the first $e$ trajectories into $k$ microbatches,
\begin{equation}
F[k,e]=\min_{\substack{s<e\\W(s+1,e)\leq T}}
\{F[k-1,s]+W(s+1,e)\}.
\label{eq:contiguous-dp}
\end{equation}
Set $F[0,0]=0$ and infeasible states to $+\infty$.
Only feasible predecessor states participate. Among equal-total-work choices,
the planner minimizes the largest individual microbatch work. For fixed $e$,
$W(s+1,e)$ is nonincreasing as $s$ moves right, so starts before the first
capacity-feasible one can be skipped without losing a legal partition. This
exactness covers only the contiguous partition space and the count range in
Equation~\eqref{eq:k-search-range}; the exact candidates remain in the final
comparison rather than being replaced by a greedy candidate.

For $N>N_0=512$, scalable mode uses a sampled-count budget $Q_{\max}=32$.
It keeps all counts if they fit the budget; otherwise it keeps both endpoints,
dense low-count points starting at $K_{\min}$, and sparse points across the
range. Starting from $\mathcal P_0$, it reaches a smaller target $K$ by repeatedly
merging adjacent feasible groups $A,B$ with maximum saving
$C(A)+C(B)-C(A\cup B)$. For a larger $K$, it repeatedly selects a contiguous split
$M=L\cup R$ with minimum increase $C(L)+C(R)-C(M)$. Construction stops at the
target count, or fails if no legal operation remains.

\paragraph{Trie-aware greedy candidates.}

At both scales, the planner independently constructs additional candidates
without a preset count or modifications to existing candidates. If trajectory
$x$ is inserted between neighbors $p,q$ in a microbatch's lexicographic order,
its exact work increment is
\begin{equation}
\begin{aligned}
\Delta C={}&|x|-\operatorname{LCP}(p,x)-\operatorname{LCP}(x,q)\\
&+\operatorname{LCP}(p,q).
\end{aligned}
\label{eq:greedy-insertion-delta}
\end{equation}
Terms for nonexistent neighbors are omitted at either endpoint. Four processing
orders reduce order sensitivity: forward and reverse token-lexicographic order,
descending trajectory length, and descending unique-suffix score
$u(x)=|x|-\max(\operatorname{LCP}(p_x,x),\operatorname{LCP}(x,q_x))$, where
$p_x,q_x$ are $x$'s global lexicographic neighbors. The last order prioritizes
branches with fewer sharing opportunities.

Each order uses two placement rules. \emph{Best-fit} chooses the feasible target
with greatest work after insertion, reducing capacity fragmentation.
\emph{Max-shared} instead maximizes $|x|-\Delta C$. A new microbatch is created
if no existing one fits. Each of the eight resulting plans attempts to eliminate
small groups, then aligns $K$ to $D$ through singleton splits with minimum work
increase. The greedy candidate is preselected by microbatch count, then total
compact work. This construction rule does not replace the final compression-first
objective: the candidate participates even if its count lies outside the base
range in Equation~\eqref{eq:k-search-range}.

\paragraph{Initial assignment.}

For each candidate, membership and count are temporarily fixed. Microbatches
are sorted by decreasing work, and each consecutive $D$ forms a slot. Within
each slot, heavier microbatches go to replicas with lower cumulative work.
The planner then compares all feasible candidates lexicographically using
$J_{\mathrm{init}}$ in Equation~\eqref{eq:prefix-plan-objective}, selecting
the best candidate under that objective. Only the specified contiguous
search is exact; sampled evolution, greedy construction, assignment, and slot
construction use the stated heuristic rules.

\subsection{L2: Reuse-Aware Placement}
\label{supp:scheduling-l2}

L2 refines L1's partition and assignment through replica--slot placement,
sharing windows, and whole-trajectory membership changes.

\paragraph{Work and scoring.}
Let $\mathcal E(\Pi)$ index all window producers and consumer slots. For
replica $d$, $n_{d,e}$ counts tensor rows, including hardware-alignment padding,
for producer or consumer $e$; a shared consumer contributes its residual, an unshared
consumer its complete execution. Residuals retain the original microbatch's
padding before realignment. Let $i$ index individual Q/KV sequence pairs
within the full-attention packed calls for $e$, with $q_i$ queries and $k_i$
keys, including cached ancestors. The forward/backward cost estimates are
\begin{equation}
\begin{aligned}
A_{d,e}&=\sum_i\bigl[q_i(k_i-q_i)+q_i(q_i+1)/2\bigr],\\
R_{d,e}&=\sum_i q_i k_i,\\
\widehat c^{\mathrm F}_{d,e}&=\alpha_{\mathrm F}n_{d,e}
  +\beta_{\mathrm F}A_{d,e},\\
\widehat c^{\mathrm B}_{d,e}&=\alpha_{\mathrm B}n_{d,e}
  +\beta_{\mathrm B}R_{d,e}.
\end{aligned}
\label{eq:l2-phase-cost}
\end{equation}
$A_{d,e}$ counts causally visible query--key pairs; $R_{d,e}$ sums the
per-sequence query--key rectangle areas. The coefficients are fixed
model/backend parameters. These costs
enter the separate forward/backward DP maxima in
Equation~\eqref{eq:phase-placement-objective}. Complete plans are compared
lexicographically using
\begin{equation}
\begin{aligned}
N_d&=\sum_{e\in\mathcal E(\Pi)}n_{d,e},\qquad
W_{\mathrm{phys}}=\sum_d N_d,\\
B_{\mathrm{DP}}&=
\frac{\sum_e(\max_d n_{d,e}-\min_d n_{d,e})}
     {\max(1,\sum_e\max_d n_{d,e})},\\
J_{\mathrm{L2}}(\Pi)&=\bigl(\Phi(\Pi),W_{\mathrm{phys}},
B_{\mathrm{DP}},\max_d N_d\bigr).
\end{aligned}
\label{eq:l2-plan-score}
\end{equation}
Window-local refinement uses the same first two terms, followed by
the unnormalized sum of per-computation row-count ranges and the maximum
per-replica row total. The model
does not directly estimate communication, linear-attention replay, or pipeline
overlap.
L2 selects the best-scoring feasible plan among the candidates examined.

\paragraph{Candidate layouts and windows.}
Three candidate layouts are examined: L1's balanced assignment, token-lexicographic
placement, and prefix-affinity placement. Affinity candidates use a bounded
neighborhood and greedily select disjoint groups by the compact-token savings
from merging their tries. Each layout is evaluated
with several window-length caps. Consecutive windows are scanned longest
first; all replicas must admit feasible producer forests at common boundaries.
For replica $d$ and window $w$, let $h_{d,j}$ count cached compact tokens,
deduplicated within consumer $j$, and $p_{d,w}$ count compact tokens computed
by the producer. Net decoder-token savings must satisfy
\begin{equation}
g_{d,w}=\sum_{j\in w}h_{d,j}-p_{d,w}>0.
\label{eq:l2-net-savings}
\end{equation}
\paragraph{Local refinement.}
Within each window, intact trajectories are repacked in descending estimated
residual cost into the existing consumer count. Exact LCP costs guide packing;
capacity checks cover full consumer layouts and reconstructed producers.
Consumers are ordered by descending residual forward-plus-backward
cost across replicas. Coordinated and replica-local depth adjustments
reconstruct producer and residual work. Bounded whole-trajectory exchanges
target heavily and lightly loaded replicas at corresponding slots, with
repacking where feasible. An exchange must improve the local score without
increasing the producer/residual token count including alignment padding.

%% file: appendices/b_minimum_call.tex
\section{Minimum-Call Plan Recovery}
\label{supp:minimum-call}

The execution model and recurrences are given in
Section~\ref{sec:call-planning}. A segment is a maximal
fork-free token path, not a fixed-length chunk. A packed call may process many
independent variable-length sequences, each spanning multiple chunks. Its
internally produced state becomes an independent sequence's initial state only
after return. The planner fixes nearest-anchor replay and produces every
non-replay output once; it does not use earlier-anchor multichunk replay or
duplicate segment computation to reduce call depth.

\begin{algorithm*}[t]
\caption{Minimum-call planning and plan recovery}
\label{alg:minimum-call}
\begin{algorithmic}[1]
\REQUIRE Rooted segment tree $G$, chunk size $B$
\ENSURE Ordered rounds of packed sequences and requested states
\FOR{each segment $v$ in postorder}
  \IF{$v$ is a leaf}
    \STATE $T_{\mathrm{pre}}[v]\gets T_{\mathrm{in}}[v]\gets1$
  \ELSE
    \STATE Obtain the largest and second-largest child $T_{\mathrm{pre}}$
    \FOR{each child $o$}
      \STATE $h_o\gets\max_{u\neq o}T_{\mathrm{pre}}[u]$
    \ENDFOR
    \STATE $T_{\mathrm{in}}[v]\gets\min_o\max(T_{\mathrm{in}}[o],1+h_o)$
    \STATE $T_{\mathrm{pre}}[v]\gets\max_u T_{\mathrm{pre}}[u]$ if $e_v<a_v+B$, else $T_{\mathrm{in}}[v]$
  \ENDIF
\ENDFOR
\STATE $K_{\mathrm{round}}\gets T_{\mathrm{pre}}[\mathrm{root}]$; create zero-state root sequence $q_0$ in round $1$
\FUNCTION{$\operatorname{Recover}(v,\mathrm{ready},K_{\mathrm{round}},r,q)$}
\STATE Append $v$ to $q$; \textbf{if} $v$ is a leaf \textbf{then return}
\STATE $same\gets\mathrm{ready}\wedge(e_v<a_v+B)$
\STATE $\delta\gets0$ if $same$, else $1$; $K_{\mathrm{side}}\gets K_{\mathrm{round}}-\delta$
\STATE For each child $o$: $H[o]\gets T_{\mathrm{pre}}[o]$ if $same$, else $T_{\mathrm{in}}[o]$
\STATE $f\gets\sum_u\ind[T_{\mathrm{pre}}[u]>K_{\mathrm{side}}]$
\STATE Choose the first child $o$ such that
\STATE \hspace{1em}$H[o]\leq K_{\mathrm{round}}$ and $f-\ind[T_{\mathrm{pre}}[o]>K_{\mathrm{side}}]=0$
\STATE $\operatorname{Recover}(o,same,K_{\mathrm{round}},r,q)$
\FOR{each side child $u\neq o$}
  \IF{$same$}
    \STATE Launch $q_u$ in round $r$ from $\mathbf S_{a_v}$ with fixed replay
  \ELSE
    \STATE Request $\mathbf S_{b_v}$; launch $q_u$ in round $r+1$ with fixed replay
  \ENDIF
  \STATE $\operatorname{Recover}(u,\mathrm{true},K_{\mathrm{side}},r+\delta,q_u)$
\ENDFOR
\ENDFUNCTION
\STATE $\operatorname{Recover}(\mathrm{root},\mathrm{true},K_{\mathrm{round}},1,q_0)$
\STATE Batch all sequences assigned to each round into one variable-length call; \textbf{return} rounds
\end{algorithmic}
\end{algorithm*}

\subsection{State Readiness}

For $v=[s_v,e_v)$, let $a_v=a(s_v)$ and $b_v=a(e_v)$, with
$a(p)=B\lfloor p/B\rfloor$. A side child replays $[b_v,e_v)$ from
$\mathbf S_{b_v}$. If the entry anchor was ready before the call and
$e_v<a_v+B$, then $b_v=a_v$, and the side sequence may begin in that same call.
At equality, zero replay does not imply a ready state: the boundary is produced
inside the call and can be consumed independently only in the next round.

In Figure~\ref{fig:minimum-call}, $p_1=B-1$ uses the
already-ready zero anchor, so $v_2$ can replay $[0,p_1)$ in Round~1. At
$p_2=B+1$, the first call produces $\mathbf S_B$; $v_4$ starts from that state
only in Round~2, replaying $[B,p_2)$. The direct chain
$v_0\!\rightarrow v_1\!\rightarrow v_3$ stays in Round~1. Thus the whole tree
needs two calls even though its two side branches have different state-readiness
conditions.

\subsection{Budget-Aware Plan Recovery}

Algorithm~\ref{alg:minimum-call} first computes the two minimum-round values
using Equations~\eqref{eq:tin} and~\eqref{eq:tpre}. For each node,
the largest and second-largest child $T_{\mathrm{pre}}$ values provide all
sibling maxima in a linear number of child visits.

Recovery starts with global round budget
$K_{\mathrm{round}}=T_{\mathrm{pre}}(\mathrm{root})$, distinct from the
microbatch count $K$. At a node, $\delta$ records whether side children must
wait; $K_{\mathrm{side}}=K_{\mathrm{round}}-\delta$ is their remaining budget.
The count $f$ identifies children that exceed it. The first child in fixed order
whose continuation fits and whose removal leaves no over-budget side child
becomes the direct continuation. This uses \mbox{available} slack in noncritical
subtrees rather than requiring every local choice to minimize its own depth.
Each edge is visited a constant number of times in both value computation and
recovery, giving $O(|V|+|E|)$ time and $O(|V|)$ space.

\subsection{Replay Accounting and Packed Output}

For an internal segment $v$, all but its direct child replay $e_v\bmod B$
ancestor positions. The mandatory replay total is
\begin{equation}
R=\sum_{v:\operatorname{ch}(v)\neq\varnothing}
\bigl(|\operatorname{ch}(v)|-1\bigr)(e_v\bmod B).
\label{eq:mandatory-replay}
\end{equation}
It counts repeated linear-attention state updates, not additional compact tokens
or semantic outputs. Its value is independent of the selected direct child,
so minimum-call \mbox{planning} changes sequential depth while preserving $R$.

Each recovered direct-continuation chain forms one input sequence. A side
sequence carries its fixed replay interval and enters the earliest round
allowed by state readiness. Each round packs all its sequences with explicit
initial-state sources, replay intervals, and requested boundary states. The
runtime consumes this deterministic plan without replanning the tree.
Appendix~\ref{supp:planning-cases} expands three irregular examples.

%% file: appendices/c_backend_semantics.tex
\section{Backend and Semantic Details}
\label{supp:backend-semantics}

This section expands the shared-execution and semantic-restoration interfaces
in Sections~\ref{sec:hybrid-execution} and~\ref{sec:semantic-restoration}. The design applies to full/linear-attention layers and
dense/MoE feed-forward layers. Multi-head Latent Attention (MLA) and Kimi Delta
Attention (KDA) are the concrete backends in the integrated Ling-3.0-tiny and
Ling-3.0-flash models~\citep{inclusionai2026ling3tiny,inclusionai2026ling3flash};
the general planner does not depend on their particular projection or state
update formulas.

\begin{figure*}[!t]
  \centering
  \includegraphics[trim=0 4bp 0 0,clip,width=\textwidth]{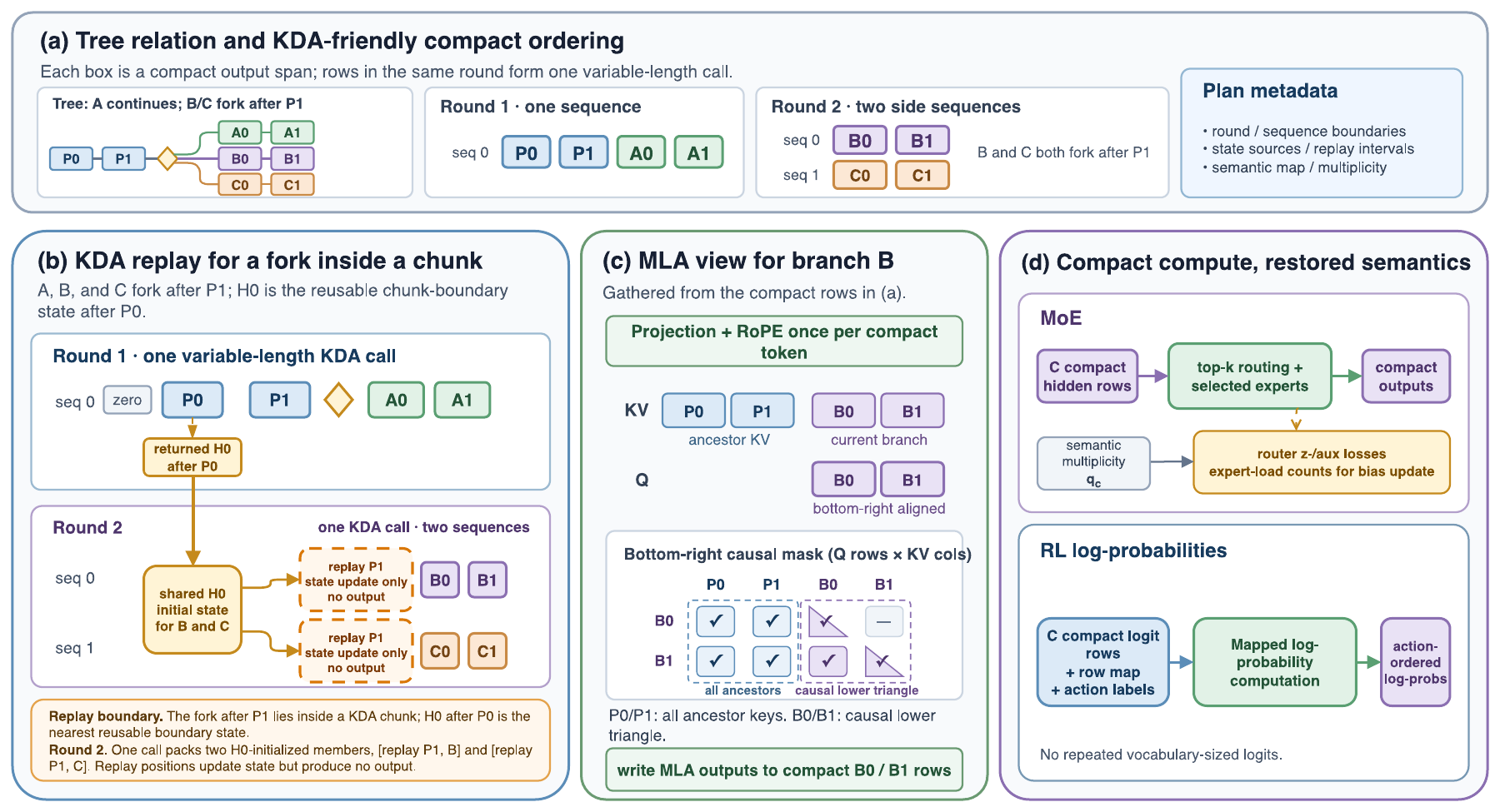}%
  \caption{\textbf{One compact physical ordering across hybrid-attention training.}
  (a) Shared path P0--P1 branches into A, B, and C; A directly continues in
  Round~1, while B/C form independent sequences in Round~2.
  (b) Both side sequences start from the boundary state $H_0$ after P0 and replay
  P1 inside the KDA core, without writing semantic outputs.
  (c) B's MLA queries read ancestor KV P0/P1. Within B, diagonal blocks are
  causal; B1 reads earlier B0, not conversely. Outputs return to compact rows.
  (d) Compact routing/expert execution uses multiplicity-weighted router losses
  and expert-load counts; mapped log-probabilities restore original token order
  without expanded vocabulary-sized logits.}
  \label{fig:hybrid-execution}
\end{figure*}

\subsection{Chunkwise State Propagation}

As one concrete example, KDA's chunk-level update~\citep{kimiLinear2025}, with
chunk length denoted by $B$ to distinguish it from compact-token count $C$, is
\begin{equation}
\begin{split}
\mathbf S_{[c+1]}={}&
\operatorname{Diag}\!\left(\gamma_{[c]}^{B}\right)\mathbf S_{[c]}\\
&+\left(\Gamma_{[c]}^{i\rightarrow B}\odot\mathbf K_{[c]}\right)^{\!\top}
\left(\mathbf U_{[c]}-\mathbf W_{[c]}\mathbf S_{[c]}\right).
\end{split}
\label{eq:chunk-state}
\end{equation}
Here $\mathbf S_{[c]}$ is the recurrent state at chunk $c$'s start;
$\mathbf K_{[c]},\allowbreak\mathbf U_{[c]},\allowbreak\mathbf W_{[c]}$, and
$\boldsymbol\Gamma_{[c]}$ are its dense-update and decay summaries.
Other backends may use different recurrences. The common requirement is
chunkwise propagation of recoverable boundary states with a differentiable
initial state.

For sequence $i$, the requested indices $\mathcal I_i$ in
Equation~\eqref{eq:chunkla-interface} select states already produced by that
propagation. The backend writes out and retains only those requested states,
with their graph connections, rather than scanning the state sequence again.
Backward adds each later sequence's initial-state gradient to the preceding
computation. Multiple readers of one boundary tensor contribute gradients to
that tensor. Packed sequences without explicit sharing remain state- and
gradient-isolated.

\subsection{Compact Ordering and Full-Attention Views}

Direct-continuation segments form contiguous input sequences in the planned
compact-token ordering. Sequences in one round form one packed call; replay sources
and requested states are directly addressed from the plan. Full attention
constructs temporary Q/KV views from this \mbox{ordering}, while MoE routes compact
tokens directly. No layer reconstructs a trajectory-expanded layout.

For MLA, projections and rotary position embeddings (RoPE) execute once on the
compact tokens required by the current computation, using original trajectory position
IDs. A branch with $h$ ancestor tokens and $\ell$ current tokens uses
\begin{equation}
|Q|=\ell,\qquad |KV|=h+\ell.
\label{eq:asymmetric-qkv}
\end{equation}
Queries contain only current tokens; keys and values contain ancestors plus
current tokens. Separate packed boundaries describe Q and KV.
Transformer Engine~\citep{nvidia2026transformerengine} supplies bottom-right
causal alignment: query $i\in\{0,\ldots,\ell-1\}$ can read exactly keys
$j\in\{0,\ldots,h+\ell-1\}$ satisfying
\begin{equation}
j\leq h+i.
\label{eq:bottom-right}
\end{equation}
Top-left alignment would incorrectly align short Q with the start of KV.
Outputs are index-copied to unique compact positions; gather backward sums
ancestor-view gradients into the corresponding shared projected rows.
Only projected ancestor KV views are temporarily expanded. Projection,
RoPE, token-wise feed-forward computation, and final outputs remain compact.
This construction uses the existing packed-attention backend, not an
arbitrary-tree custom kernel. It also applies to multi-head attention (MHA),
grouped-query attention (GQA), and other full-attention variants supporting
asymmetric packed Q/KV views.

\subsection{Replay and Bounded Local Context}

KDA consumes the planned round ordering, with one packed call per round rather
than one per tree node. Side branches gather already projected and convolved
ancestor activations for replay in the state-update core.

A causal convolution of width $w$ cannot treat neighboring packed sequences as
history. Each independent sequence receives a temporary view of the most recent
$w-1$ projected activations from its true ancestors. Segment identifiers prevent
convolution windows from crossing sequences, and each compact token retains one
post-convolution result. Replay gathers these results without reconvolving
them; indexed-gather backward accumulates all reader gradients into the original
projection/convolution graph. Thus bounded input views supply local context,
while boundary-state handoff and fixed core replay restore recurrent state.

\subsection{Window Interfaces and Resource Lifetimes}

Section~\ref{sec:cross-mb-runtime} defines the cached
interfaces and window-scoped gradient accumulation. PP transfers carry the
consumer's restored complete compact hidden-state tensor; the next stage selects
its uncached tokens for decoder execution. Router logits are likewise restored in each consumer's
own layout before its multiplicity-weighted statistics are computed.

Without activation recomputation, requested states and their graph connections
persist until backward. With recomputation, only necessary shared interfaces
and differentiable connections require window-scoped retention; each
microbatch's complete internal activations need not remain live. Reconstruction
uses the fixed plan and preserves gradient accumulation and the single producer
backward pass.

\subsection{Parallel Execution}

\harts{} composes with DP, tensor parallelism (TP), sequence parallelism
(SP)~\citep{korthikanti2023recomputation}, PP~\citep{narayanan2021megatron},
and expert parallelism (EP). DP replicas are logical scheduling units,
not process ranks; their participating parallel groups can overlap. TP keeps
the original parameter partition. SP evenly partitions compact tensor rows,
including hardware-alignment padding, within a TP group. All PP stages follow the same consumer order,
window boundaries, and compact layouts. EP routing, communication, and expert
execution operate on compact tokens.

\harts{} also supports DeepSpeed-Ulysses context parallelism
(CP)~\citep{jacobs2023deepspeedulysses}. Its all-to-all transforms sequence
partitions into independent attention-head partitions; the attention computation
remains distributed by heads, independent of the internal attention execution
used here.

For the pipeline in Figure~\ref{fig:cross-mb-pipeline},
asynchronous point-to-point transfers wait only when a received tensor is
consumed or a send buffer is reclaimed, without requiring a global barrier.

\subsection{Compact Logit Gradients and MoE Statistics}

Mapped cross-entropy returns $-\lambda_s$ from
Equation~\eqref{eq:mapped-logp}; negation recovers the current
log-probability. Under Section~\ref{sec:semantic-restoration}'s policy
interface, the chain rule gives
\begin{equation}
\nabla_{z_c}\mathcal L_{\mathrm{policy}}
=\sum_{\substack{s\in\mathcal A\\m(s)=c}}
\frac{\partial\mathcal L_{\mathrm{policy}}}{\partial\lambda_s}
\nabla_{z_c}\lambda_s.
\label{eq:mapped-logp-grad}
\end{equation}
Each semantic position contributes its own gradient to the shared row; the
implementation produces only the compact logit-gradient tensor $d\mathrm{Logits}$, without an $S\times V$
tensor.

In vocabulary-parallel execution, TP processes reduce maxima and sums of
exponentials for each compact row, plus a target-logit scalar for each
nonfinal semantic position. They do not all-gather complete vocabulary rows.

MoE statistics use the multiplicity weights from
Section~\ref{sec:semantic-restoration} under deterministic top-$k$ routing
without token dropping. Dispatch, EP communication, selected-expert computation,
and combine remain compact. Hardware-alignment padding contributes no semantic
positions and does not re-expand shared tokens.

%% file: appendices/d_planning_cases.tex
\section{Linear-Attention Execution Planning Cases}
\label{supp:planning-cases}
\label{appendix:linear-attention-case}

These three irregular trees illustrate Algorithm~\ref{alg:minimum-call}, all
with chunk size $B=64$. A new independent sequence resumes from the preceding
64-token boundary and replays to its branch entry when needed. Circles list
segment names, token counts, and calls; colors identify calls. Thick edges
form direct-continuation chains. Dashed edges start independent sequences,
with labels $r=k$ denoting replay counts; zero labels are omitted.
The totals $R$ follow Equation~\eqref{eq:mandatory-replay}.

\begin{figure}[!ht]
  \centering
  \includegraphics[width=0.96\textwidth,height=0.62\textheight,keepaspectratio]{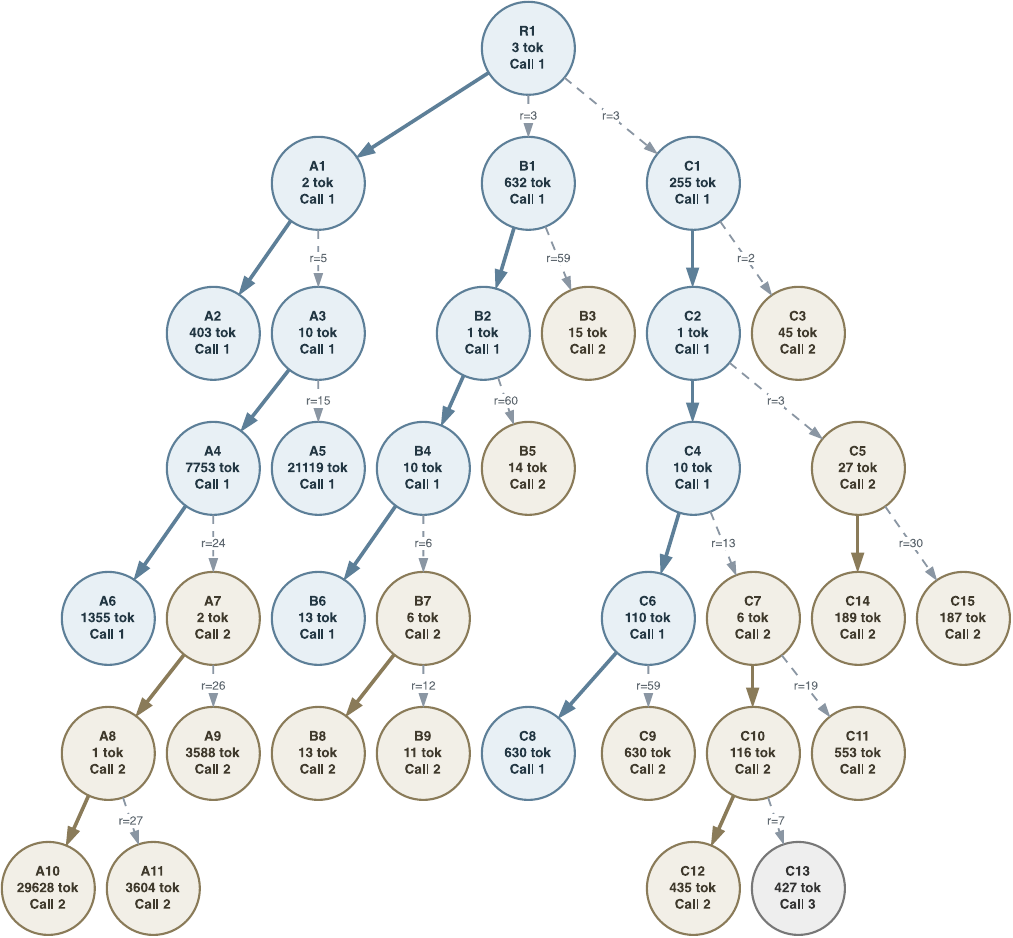}%
  \caption{\textbf{Case I: three calls and $R=373$.} Circles show per-segment
  token counts. Thick solid edges form direct-continuation chains, while dashed
  edges start independent sequences and are labeled with their replay length
  under chunk size $B=64$.}
  \label{fig:appendix-w08}
  \vspace{0.4em}
  \begin{minipage}[t]{0.485\textwidth}\footnotesize\raggedright
    \textbf{Call 1 (five sequences; replay 26).}
    $R1\!\rightarrow A1\!\rightarrow A2$ ($r=0$);
    $A3\!\rightarrow A4\!\rightarrow A6$ (5); $A5$ (15);
    $B1\!\rightarrow B2\!\rightarrow B4\!\rightarrow B6$ (3); and
    $C1\!\rightarrow C2\!\rightarrow C4\!\rightarrow C6\!\rightarrow C8$ (3).
  \end{minipage}\hfill
  \begin{minipage}[t]{0.485\textwidth}\footnotesize\raggedright
    \textbf{Calls 2--3.} Call~2 has thirteen sequences:
    $A7\!\rightarrow A8\!\rightarrow A10$ (24), $A11$ (27), $A9$ (26),
    $B7\!\rightarrow B8$ (6), $B9$ (12), $B5$ (60), $B3$ (59),
    $C9$ (59), $C7\!\rightarrow C10\!\rightarrow C12$ (13), $C11$ (19),
    $C5\!\rightarrow C14$ (3), $C15$ (30), and $C3$ (2), totaling 340 replay tokens.
    Call~3 contains only $C13$ ($r=7$).
  \end{minipage}
\end{figure}

\clearpage
\begin{figure}[p]
  \centering
  \includegraphics[width=0.96\textwidth,height=0.80\textheight,keepaspectratio]{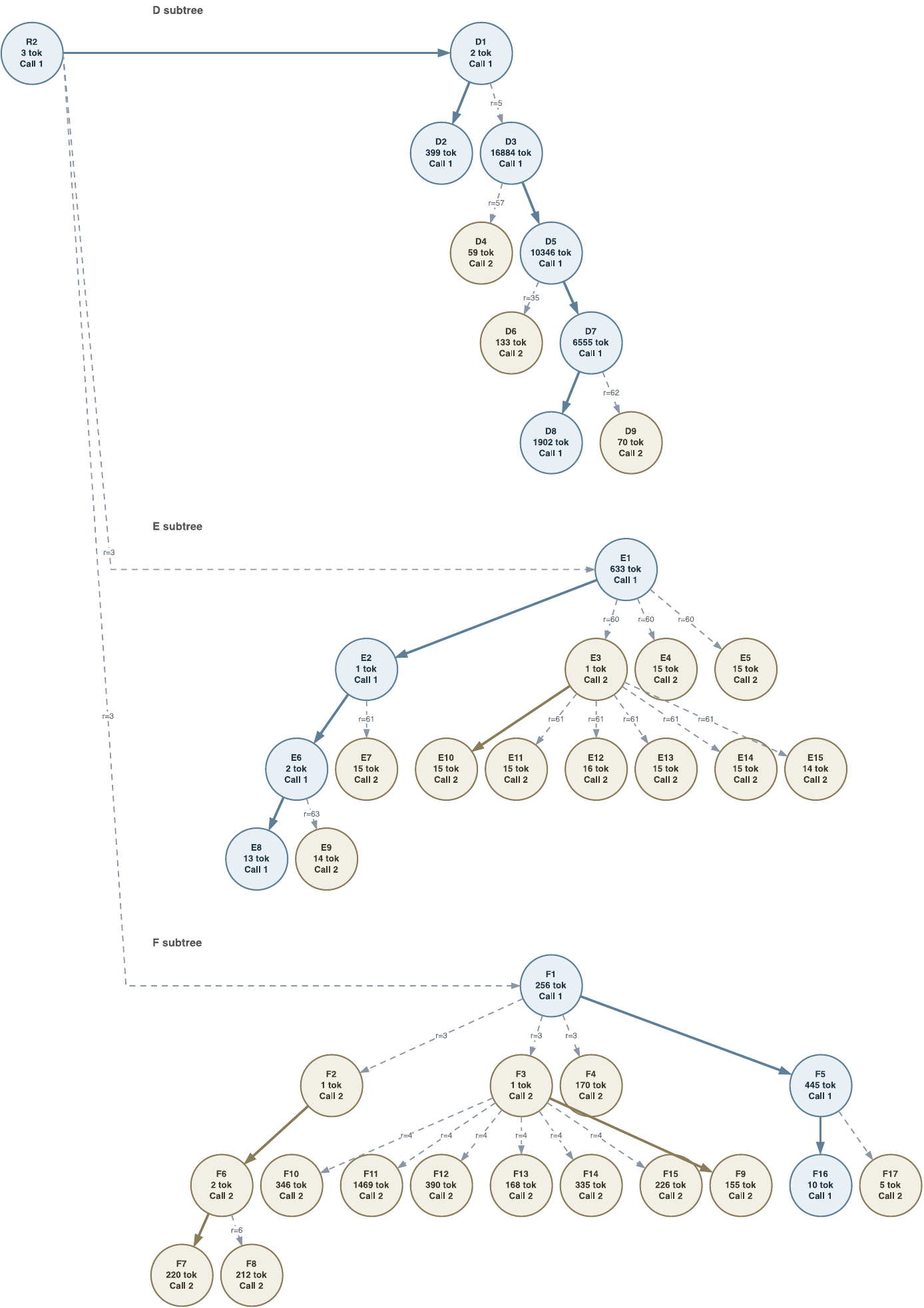}%
  \caption{\textbf{Case II: two calls and $R=813$.} The three child subtrees of
  the single shared root $R2$ are unfolded vertically only to preserve label
  readability. Under chunk size $B=64$, replay labels expose how many branches
  become ready together for the second variable-length call.}
  \label{fig:appendix-m05}
  \vspace{0.4em}
  \begin{minipage}[t]{0.485\textwidth}\footnotesize\raggedright
    \textbf{Call 1 (four sequences; replay 11).}
    $R2\!\rightarrow D1\!\rightarrow D2$ ($r=0$);
    $D3\!\rightarrow D5\!\rightarrow D7\!\rightarrow D8$ (5);
    $E1\!\rightarrow E2\!\rightarrow E6\!\rightarrow E8$ (3); and
    $F1\!\rightarrow F5\!\rightarrow F16$ (3).
  \end{minipage}\hfill
  \begin{minipage}[t]{0.485\textwidth}\footnotesize\raggedright
    \textbf{Call 2 (24 sequences; replay 802).}
    $D4,D6,D9$ replay $57,35,62$.
    On the $E$ side: $E9$ (63), $E7$ (61), $E3\!\rightarrow E10$ (60),
    $E11$--$E15$ (61 each), and $E4,E5$ (60 each).
    On the $F$ side: $F2\!\rightarrow F6\!\rightarrow F7$ (3), $F8$ (6),
    $F3\!\rightarrow F9$ (3), $F10$--$F15$ (4 each), $F4$ (3), and $F17$ (0).
  \end{minipage}
\end{figure}

\clearpage
\begin{figure}[p]
  \centering
  \includegraphics[width=0.96\textwidth,height=0.80\textheight,keepaspectratio]{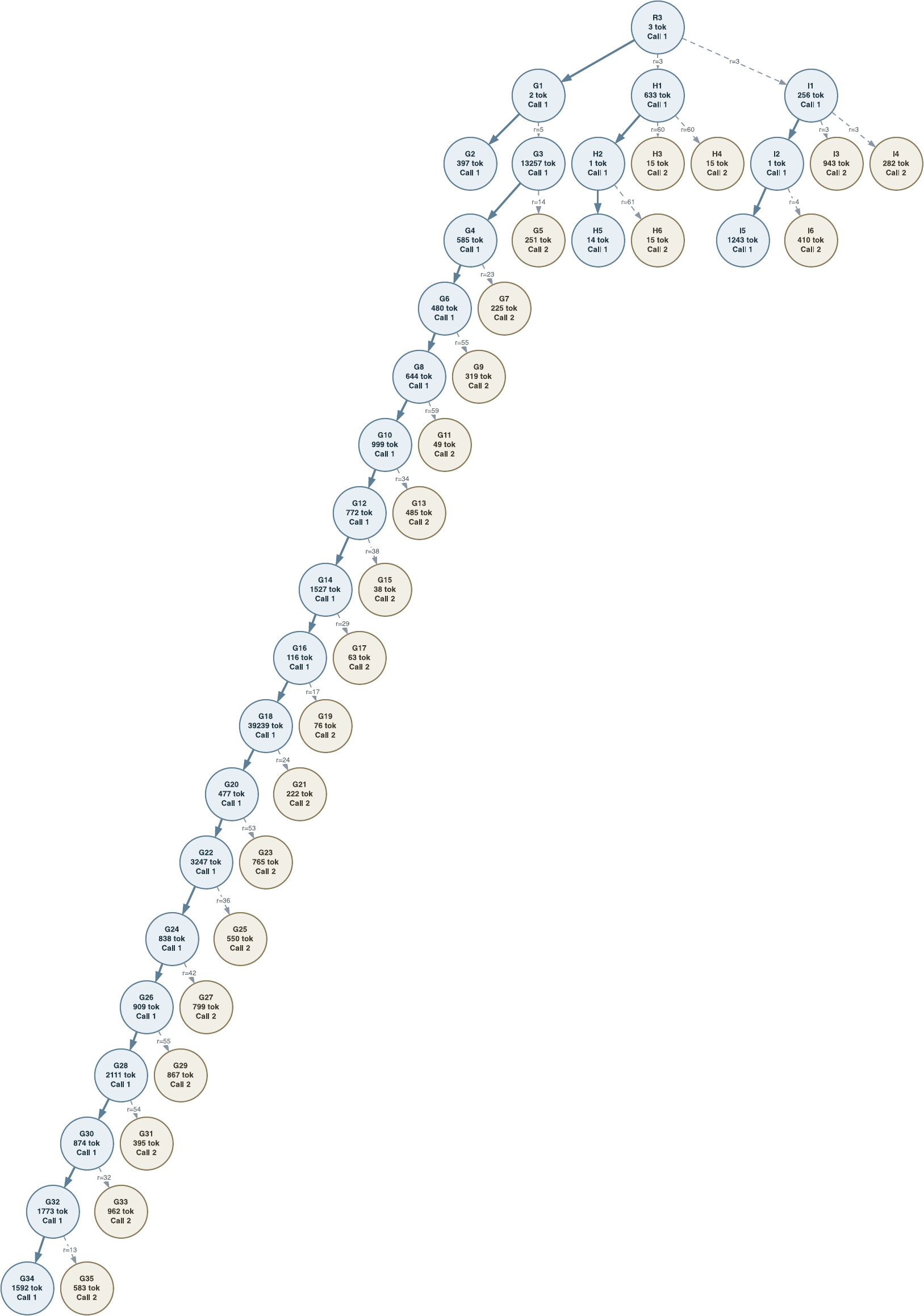}%
  \caption{\textbf{Case III: two calls and $R=780$.} With chunk size $B=64$,
  the first call follows the long $G$ continuation chain while also executing
  three replay-started sequences; all side leaves become ready for one parallel
  second call.}
  \label{fig:appendix-m12}
  \vspace{0.4em}
  \begin{minipage}[t]{0.485\textwidth}\footnotesize\raggedright
    \textbf{Call 1 (four sequences; replay 11).}
    $R3\!\rightarrow G1\!\rightarrow G2$ ($r=0$);
    $G3\!\rightarrow G4\!\rightarrow G6\!\rightarrow\cdots\rightarrow G32\!\rightarrow G34$ (5);
    $H1\!\rightarrow H2\!\rightarrow H5$ (3); and
    $I1\!\rightarrow I2\!\rightarrow I5$ (3).
  \end{minipage}\hfill
  \begin{minipage}[t]{0.485\textwidth}\footnotesize\raggedright
    \textbf{Call 2 (22 single-segment sequences; replay 769).}
    $G35,G33,G31,G29$ replay $13,32,54,55$;
    $G27,G25,G23,G21$ replay $42,36,53,24$;
    $G19,G17,G15,G13$ replay $17,29,38,34$; and
    $G11,G9,G7,G5$ replay $59,55,23,14$.
    The remaining $H6,H3,H4,I6,I3,I4$ replay $61,60,60,4,3,3$.
  \end{minipage}
\end{figure}

%% file: references.bib
@misc{perron2025ortools,
  title        = {{OR-Tools}},
  author       = {Laurent Perron and Vincent Furnon},
  organization = {Google},
  year         = {2025},
  note         = {Version 9.14},
  url          = {https://developers.google.com/optimization/}
}

@misc{narayanan2021megatron,
  title         = {Efficient Large-Scale Language Model Training on {GPU} Clusters Using {Megatron-LM}},
  author        = {Deepak Narayanan and Mohammad Shoeybi and Jared Casper and Patrick LeGresley and Mostofa Patwary and Vijay Anand Korthikanti and Dmitri Vainbrand and Prethvi Kashinkunti and Julie Bernauer and Bryan Catanzaro and Amar Phanishayee and Matei Zaharia},
  year          = {2021},
  eprint        = {2104.04473},
  archiveprefix = {arXiv},
  primaryclass  = {cs.CL},
  url           = {https://arxiv.org/abs/2104.04473}
}

@inproceedings{korthikanti2023recomputation,
  title     = {Reducing Activation Recomputation in Large Transformer Models},
  author    = {Vijay Anand Korthikanti and Jared Casper and Sangkug Lym and Lawrence McAfee and Michael Andersch and Mohammad Shoeybi and Bryan Catanzaro},
  booktitle = {Proceedings of Machine Learning and Systems},
  volume    = {5},
  pages     = {341--353},
  year      = {2023}
}

@misc{deepseek2024v2,
  title         = {{DeepSeek-V2}: A Strong, Economical, and Efficient Mixture-of-Experts Language Model},
  author        = {{DeepSeek-AI} and Aixin Liu and Bei Feng and Bin Wang and
                   Bingxuan Wang and Bo Liu and Chenggang Zhao and Chengqi Dengr and
                   Chong Ruan and Damai Dai and Daya Guo and Dejian Yang and
                   Deli Chen and Dongjie Ji and Erhang Li and Fangyun Lin and
                   Fuli Luo and Guangbo Hao and Guanting Chen and Guowei Li and
                   H. Zhang and Hanwei Xu and Hao Yang and Haowei Zhang and
                   Honghui Ding and Huajian Xin and Huazuo Gao and Hui Li and
                   Hui Qu and J. L. Cai and Jian Liang and Jianzhong Guo and
                   Jiaqi Ni and Jiashi Li and Jin Chen and Jingyang Yuan and
                   Junjie Qiu and Junxiao Song and Kai Dong and Kaige Gao and
                   Kang Guan and Lean Wang and Lecong Zhang and Lei Xu and
                   Leyi Xia and Liang Zhao and Liyue Zhang and Meng Li and
                   Miaojun Wang and Mingchuan Zhang and Minghua Zhang and Minghui Tang and
                   Mingming Li and Ning Tian and Panpan Huang and Peiyi Wang and
                   Peng Zhang and Qihao Zhu and Qinyu Chen and Qiushi Du and
                   R. J. Chen and R. L. Jin and Ruiqi Ge and Ruizhe Pan and
                   Runxin Xu and Ruyi Chen and S. S. Li and Shanghao Lu and
                   Shangyan Zhou and Shanhuang Chen and Shaoqing Wu and Shengfeng Ye and
                   Shirong Ma and Shiyu Wang and Shuang Zhou and Shuiping Yu and
                   Shunfeng Zhou and Size Zheng and T. Wang and Tian Pei and
                   Tian Yuan and Tianyu Sun and W. L. Xiao and Wangding Zeng and
                   Wei An and Wen Liu and Wenfeng Liang and Wenjun Gao and
                   Wentao Zhang and X. Q. Li and Xiangyue Jin and Xianzu Wang and
                   Xiao Bi and Xiaodong Liu and Xiaohan Wang and Xiaojin Shen and
                   Xiaokang Chen and Xiaosha Chen and Xiaotao Nie and Xiaowen Sun and
                   Xiaoxiang Wang and Xin Liu and Xin Xie and Xingkai Yu and
                   Xinnan Song and Xinyi Zhou and Xinyu Yang and Xuan Lu and
                   Xuecheng Su and Y. Wu and Y. K. Li and Y. X. Wei and
                   Y. X. Zhu and Yanhong Xu and Yanping Huang and Yao Li and
                   Yao Zhao and Yaofeng Sun and Yaohui Li and Yaohui Wang and
                   Yi Zheng and Yichao Zhang and Yiliang Xiong and Yilong Zhao and
                   Ying He and Ying Tang and Yishi Piao and Yixin Dong and
                   Yixuan Tan and Yiyuan Liu and Yongji Wang and Yongqiang Guo and
                   Yuchen Zhu and Yuduan Wang and Yuheng Zou and Yukun Zha and
                   Yunxian Ma and Yuting Yan and Yuxiang You and Yuxuan Liu and
                   Z. Z. Ren and Zehui Ren and Zhangli Sha and Zhe Fu and
                   Zhen Huang and Zhen Zhang and Zhenda Xie and Zhewen Hao and
                   Zhihong Shao and Zhiniu Wen and Zhipeng Xu and Zhongyu Zhang and
                   Zhuoshu Li and Zihan Wang and Zihui Gu and Zilin Li and Ziwei Xie},
  year          = {2024},
  eprint        = {2405.04434},
  archiveprefix = {arXiv},
  primaryclass  = {cs.CL},
  url           = {https://arxiv.org/abs/2405.04434}
}

@misc{zoph2022stmoe,
  title         = {{ST-MoE}: Designing Stable and Transferable Sparse Expert Models},
  author        = {Barret Zoph and Irwan Bello and Sameer Kumar and Nan Du and Yanping Huang and Jeff Dean and Noam Shazeer and William Fedus},
  year          = {2022},
  eprint        = {2202.08906},
  archiveprefix = {arXiv},
  primaryclass  = {cs.CL},
  url           = {https://arxiv.org/abs/2202.08906}
}

@misc{wang2024lossfree,
  title         = {Auxiliary-Loss-Free Load Balancing Strategy for Mixture-of-Experts},
  author        = {Lean Wang and Huazuo Gao and Chenggang Zhao and Xu Sun and Damai Dai},
  year          = {2024},
  eprint        = {2408.15664},
  archiveprefix = {arXiv},
  primaryclass  = {cs.LG},
  url           = {https://arxiv.org/abs/2408.15664}
}

@misc{jacobs2023deepspeedulysses,
  title         = {{DeepSpeed Ulysses}: System Optimizations for Enabling Training of Extreme Long Sequence Transformer Models},
  author        = {Sam Ade Jacobs and Masahiro Tanaka and Chengming Zhang and Minjia Zhang and Shuaiwen Leon Song and Samyam Rajbhandari and Yuxiong He},
  year          = {2023},
  eprint        = {2309.14509},
  archiveprefix = {arXiv},
  primaryclass  = {cs.LG},
  url           = {https://arxiv.org/abs/2309.14509}
}

@misc{wang2026treetraining,
  title         = {{Tree Training}: Accelerating Agentic {LLMs} Training via Shared Prefix Reuse},
  author        = {Jinghui Wang and Shaojie Wang and Yinghan Cui and Xuxing Chen and Chao Wang and Liang Huang and Can Tang and Xiaojiang Zhang and Junyi Peng and Li Wan and Haotian Zhang and Bin Chen},
  year          = {2026},
  eprint        = {2511.00413},
  archiveprefix = {arXiv},
  primaryclass  = {cs.LG},
  url           = {https://arxiv.org/abs/2511.00413}
}

@misc{yang2025qwen3,
  title         = {{Qwen3} Technical Report},
  author        = {An Yang and Anfeng Li and Baosong Yang and Beichen Zhang and Binyuan Hui and Bo Zheng and Bowen Yu and Chang Gao and Chengen Huang and Chenxu Lv and Chujie Zheng and Dayiheng Liu and Fan Zhou and Fei Huang and Feng Hu and Hao Ge and Haoran Wei and Huan Lin and Jialong Tang and Jian Yang and Jianhong Tu and Jianwei Zhang and Jianxin Yang and Jiaxi Yang and Jing Zhou and Jingren Zhou and Junyang Lin and Kai Dang and Keqin Bao and Kexin Yang and Le Yu and Lianghao Deng and Mei Li and Mingfeng Xue and Mingze Li and Pei Zhang and Peng Wang and Qin Zhu and Rui Men and Ruize Gao and Shixuan Liu and Shuang Luo and Tianhao Li and Tianyi Tang and Wenbiao Yin and Xingzhang Ren and Xinyu Wang and Xinyu Zhang and Xuancheng Ren and Yang Fan and Yang Su and Yichang Zhang and Yinger Zhang and Yu Wan and Yuqiong Liu and Zekun Wang and Zeyu Cui and Zhenru Zhang and Zhipeng Zhou and Zihan Qiu},
  year          = {2025},
  eprint        = {2505.09388},
  archiveprefix = {arXiv},
  primaryclass  = {cs.CL},
  url           = {https://arxiv.org/abs/2505.09388}
}

@misc{zhang2026arealdta,
  title         = {{AReaL-DTA}: Dynamic Tree Attention for Efficient Reinforcement Learning of Large Language Models},
  author        = {Jiarui Zhang and Yuchen Yang and Ran Yan and Zhiyu Mei and Liyuan Zhang and Daifeng Li and Wei Fu and Jiaxuan Gao and Shusheng Xu and Yi Wu and Binhang Yuan},
  year          = {2026},
  eprint        = {2602.00482},
  archiveprefix = {arXiv},
  primaryclass  = {cs.LG},
  url           = {https://arxiv.org/abs/2602.00482}
}

@misc{li2026scheduleprefixreuse,
  title         = {Schedule-Level Shared-Prefix Reuse for {LLM RL} Training},
  author        = {Pengbo Li and Feiyuan Zhang and Guangming Sheng and Guangxin He and Di Chai and Ziniu Li and Taiqiang Wu and Wenyu Mao and Binhang Yuan and Kai Chen},
  year          = {2026},
  eprint        = {2606.01143v3},
  archiveprefix = {arXiv},
  primaryclass  = {cs.DC},
  url           = {https://arxiv.org/abs/2606.01143v3}
}

@misc{yu2026psrl,
  title         = {{psRL}: Efficient Training for Agentic {AI} via Training-Time Prefix Sharing},
  author        = {Mianjie Yu and Zizhao Mo and Huanyu Qu and Zhirong Qian and Huanle Xu and Cen Li and Zifeng Zhao and Zhi Zhou and Jinhua Zhou and Jun Xie and Chengzhong Xu},
  year          = {2026},
  eprint        = {2608.25683},
  archiveprefix = {arXiv},
  primaryclass  = {cs.DC},
  url           = {https://arxiv.org/abs/2608.25683}
}

@misc{wang2024dpoprefixsharing,
  title         = {Accelerating Direct Preference Optimization with Prefix Sharing},
  author        = {Franklin Wang and Sumanth Hegde},
  year          = {2024},
  eprint        = {2410.20305},
  archiveprefix = {arXiv},
  primaryclass  = {cs.LG},
  url           = {https://arxiv.org/abs/2410.20305}
}

@misc{liu2025prefixgrouper,
  title         = {{Prefix Grouper}: Efficient {GRPO} Training through Shared-Prefix Forward},
  author        = {Zikang Liu and Tongtian Yue and Yepeng Tang and Longteng Guo and Junxian Cai and Qingbin Liu and Xi Chen and Jing Liu},
  year          = {2025},
  eprint        = {2506.05433},
  archiveprefix = {arXiv},
  primaryclass  = {cs.LG},
  url           = {https://arxiv.org/abs/2506.05433}
}

@misc{gai2026dualkv,
  title         = {{DualKV}: Shared-Prompt Flash Attention for Efficient {RL} Training with Large Rollouts and Long Contexts},
  author        = {Jiading Gai and Shuai Zhang and Xiang Song and Bernie Wang and George Karypis},
  year          = {2026},
  eprint        = {2605.15422},
  archiveprefix = {arXiv},
  primaryclass  = {cs.LG},
  url           = {https://arxiv.org/abs/2605.15422}
}

@misc{wang2026qwengyre,
  title         = {{QwenGyre}: An Elastic Reinforcement Learning Framework for Training {xLong-Horizon} Agents},
  author        = {Weiqi Wang and Yuxin Zhou and Mouxiang Chen and Siyuan Zhang and Yi Zhang and Yuyan Luo and Zhiyu Yin and Chencan Wu and Jiemin Jiang and Wentao Yao and Chujie Zheng and JianWei Zhang},
  year          = {2026},
  eprint        = {2609.33848v1},
  archiveprefix = {arXiv},
  primaryclass  = {cs.LG},
  url           = {https://arxiv.org/abs/2609.33848v1}
}

@inproceedings{jimenez2024swebench,
  title     = {{SWE}-bench: Can Language Models Resolve Real-world {GitHub} Issues?},
  author    = {Carlos E. Jimenez and John Yang and Alexander Wettig and Shunyu Yao and Kexin Pei and Ofir Press and Karthik R. Narasimhan},
  booktitle = {The Twelfth International Conference on Learning Representations},
  year      = {2024},
  publisher = {OpenReview.net},
  address   = {Vienna, Austria},
  numpages  = {21},
  url       = {https://openreview.net/forum?id=VTF8yNQM66}
}

@misc{shi2026tauknowledge,
  title         = {{$\tau$-Knowledge}: Evaluating Conversational Agents over Unstructured Knowledge},
  author        = {Quan Shi and Alexandra Zytek and Pedram Razavi and Karthik Narasimhan and Victor Barres},
  year          = {2026},
  eprint        = {2603.04370},
  archiveprefix = {arXiv},
  primaryclass  = {cs.AI},
  url           = {https://arxiv.org/abs/2603.04370}
}

@misc{kimiLinear2025,
  title         = {{Kimi Linear}: An Expressive, Efficient Attention Architecture},
  author        = {{Kimi Team} and Yu Zhang and Zongyu Lin and Xingcheng Yao and
                   Jiaxi Hu and Fanqing Meng and Chengyin Liu and Xin Men and
                   Songlin Yang and Zhiyuan Li and Wentao Li and Enzhe Lu and
                   Weizhou Liu and Yanru Chen and Weixin Xu and Longhui Yu and
                   Yejie Wang and Yu Fan and Longguang Zhong and Enming Yuan and
                   Dehao Zhang and Yizhi Zhang and T. Y. Liu and Haiming Wang and
                   Shengjun Fang and Weiran He and Shaowei Liu and Yiwei Li and
                   Jianlin Su and Jiezhong Qiu and Bo Pang and Junjie Yan and
                   Zhejun Jiang and Weixiao Huang and Bohong Yin and Jiacheng You and
                   Chu Wei and Zhengtao Wang and Chao Hong and Yutian Chen and
                   Guanduo Chen and Yucheng Wang and Huabin Zheng and Feng Wang and
                   Yibo Liu and Mengnan Dong and Zheng Zhang and Siyuan Pan and
                   Wenhao Wu and Yuhao Wu and Longyu Guan and Jiawen Tao and
                   Guohong Fu and Xinran Xu and Yuzhi Wang and Guokun Lai and
                   Yuxin Wu and Xinyu Zhou and Zhilin Yang and Yulun Du},
  year          = {2025},
  eprint        = {2510.26692},
  archiveprefix = {arXiv},
  primaryclass  = {cs.CL},
  url           = {https://arxiv.org/abs/2510.26692}
}

@misc{sun2023retnet,
  title         = {Retentive Network: A Successor to Transformer for Large Language Models},
  author        = {Yutao Sun and Li Dong and Shaohan Huang and Shuming Ma and Yuqing Xia and Jilong Xue and Jianyong Wang and Furu Wei},
  year          = {2023},
  eprint        = {2307.08621},
  archiveprefix = {arXiv},
  primaryclass  = {cs.CL},
  url           = {https://arxiv.org/abs/2307.08621}
}

@inproceedings{yang2024gla,
  title     = {Gated Linear Attention Transformers with Hardware-Efficient Training},
  author    = {Songlin Yang and Bailin Wang and Yikang Shen and Rameswar Panda and Yoon Kim},
  booktitle = {Proceedings of the 41st International Conference on Machine Learning},
  volume    = {235},
  year      = {2024},
  pages     = {56501--56523},
  publisher = {PMLR},
  address   = {Vienna, Austria}
}

@inproceedings{yang2024deltanet,
  title     = {Parallelizing Linear Transformers with the Delta Rule over Sequence Length},
  author    = {Songlin Yang and Bailin Wang and Yu Zhang and Yikang Shen and Yoon Kim},
  booktitle = {Advances in Neural Information Processing Systems},
  volume    = {37},
  pages     = {115491--115522},
  year      = {2024},
  publisher = {Curran Associates, Inc.},
  address   = {Vancouver, Canada},
  doi       = {10.52202/079017-3668}
}

@inproceedings{yang2025gateddeltanet,
  title     = {Gated Delta Networks: Improving {Mamba2} with Delta Rule},
  author    = {Songlin Yang and Jan Kautz and Ali Hatamizadeh},
  booktitle = {The Thirteenth International Conference on Learning Representations},
  year      = {2025},
  publisher = {OpenReview.net},
  address   = {Singapore},
  numpages  = {18}
}

@misc{shoeybi2019megatron,
  title         = {{Megatron-LM}: Training Multi-Billion Parameter Language Models Using Model Parallelism},
  author        = {Mohammad Shoeybi and Mostofa Patwary and Raul Puri and Patrick LeGresley and Jared Casper and Bryan Catanzaro},
  year          = {2019},
  eprint        = {1909.08053},
  archiveprefix = {arXiv},
  primaryclass  = {cs.CL},
  url           = {https://arxiv.org/abs/1909.08053}
}

@misc{nvidia2026transformerengine,
  title        = {{NVIDIA Transformer Engine}},
  author       = {{NVIDIA Corporation}},
  year         = {2026},
  howpublished = {Software library},
  url          = {https://github.com/NVIDIA/TransformerEngine},
  urldate      = {2026-08-23}
}

@misc{inclusionai2026ling3tiny,
  title        = {{Ling-3.0-tiny}},
  author       = {{InclusionAI}},
  year         = {2026},
  howpublished = {Hugging Face model card},
  url          = {https://huggingface.co/inclusionAI/Ling-3.0-tiny},
  urldate      = {2026-08-23}
}

@misc{inclusionai2026ling3flash,
  title        = {{Ling-3.0-flash}},
  author       = {{InclusionAI}},
  year         = {2026},
  howpublished = {Hugging Face model card},
  url          = {https://huggingface.co/inclusionAI/Ling-3.0-flash},
  urldate      = {2026-10-22}
}

@misc{shao2024deepseekmath,
  title         = {{DeepSeekMath}: Pushing the Limits of Mathematical Reasoning in Open Language Models},
  author        = {Zhihong Shao and Peiyi Wang and Qihao Zhu and Runxin Xu and Junxiao Song and Xiao Bi and Haowei Zhang and Mingchuan Zhang and Y. K. Li and Y. Wu and Daya Guo},
  year          = {2024},
  eprint        = {2402.03300},
  archiveprefix = {arXiv},
  primaryclass  = {cs.CL},
  url           = {https://arxiv.org/abs/2402.03300}
}

@misc{schulman2017ppo,
  title         = {Proximal Policy Optimization Algorithms},
  author        = {John Schulman and Filip Wolski and Prafulla Dhariwal and Alec Radford and Oleg Klimov},
  year          = {2017},
  eprint        = {1707.06347},
  archiveprefix = {arXiv},
  primaryclass  = {cs.LG},
  url           = {https://arxiv.org/abs/1707.06347}
}

@inproceedings{fu2025areal,
  title     = {{AReaL}: A Large-Scale Asynchronous Reinforcement Learning System for Language Reasoning},
  author    = {Wei Fu and Jiaxuan Gao and Xujie Shen and Chen Zhu and Zhiyu Mei and Chuyi He and Shusheng Xu and Guo Wei and Jun Mei and Jiashu Wang and Tongkai Yang and Binhang Yuan and Yi Wu},
  booktitle = {Advances in Neural Information Processing Systems},
  volume    = {38},
  pages     = {36256--36282},
  year      = {2025},
  doi       = {10.52202/085713-1218}
}
